\documentclass{article} 
\usepackage{iclr2025_conference,times}
\iclrfinalcopy

\usepackage{amsmath,amsfonts,bm}

\def\eqref#1{equation~\ref{#1}}

\def\1{\bm{1}}

\DeclareMathAlphabet{\mathsfit}{\encodingdefault}{\sfdefault}{m}{sl}
\SetMathAlphabet{\mathsfit}{bold}{\encodingdefault}{\sfdefault}{bx}{n}

\usepackage{hyperref}
\hypersetup{
    colorlinks=true,
    linkcolor=perfred,     
    citecolor=perfblue,     
    filecolor=magenta,  
    urlcolor=perfred,      
    pdfborder={0 0 0}   
}

\usepackage{url}

\usepackage[utf8]{inputenc} 
\usepackage[T1]{fontenc}    
\usepackage{url}            
\usepackage{booktabs}       
\usepackage{amsfonts}       
\usepackage{nicefrac}       
\usepackage{microtype}      
\usepackage{microtype}
\usepackage{graphicx}
\usepackage{subfigure}
\usepackage{tabularx}
\usepackage{wrapfig}
\usepackage{array}
\usepackage{microtype}
\usepackage{colortbl}      

\usepackage{caption}
\usepackage{amsfonts}       
\usepackage{amsmath}
\usepackage{amssymb}
\usepackage{mathtools}
\usepackage{amsthm}
\usepackage{bbm}
\usepackage{lipsum}

\usepackage{algorithm}
\usepackage{algpseudocode}
\usepackage{adjustbox}
\usepackage[font=small]{caption}

\usepackage{fancyvrb}
\usepackage{float}         
\usepackage{enumitem}
\usepackage{multirow}
\usepackage{rotating}
\usepackage{xspace}
\usepackage{soul}
\usepackage[toc,page,header]{appendix}
\usepackage{minitoc}

\usepackage{listings}
\usepackage{float}

\definecolor{codegray}{rgb}{0.5,0.5,0.5}
\definecolor{codepurple}{rgb}{0.58,0,0.82}
\definecolor{backcolour}{rgb}{0.95,0.95,0.92}
\lstdefinestyle{mystyle}{
    backgroundcolor=\color{white},       
    commentstyle=\color{codegray}\itshape, 
    keywordstyle=\color{blue}\bfseries,  
    numberstyle=\tiny\color{gray},       
    stringstyle=\color{codepurple},      
    basicstyle=\ttfamily\footnotesize,   
    breakatwhitespace=false,             
    breaklines=true,                     
    captionpos=b,                        
    keepspaces=true,                     
    numbersep=5pt,                       
    showspaces=false,                    
    showstringspaces=false,              
    showtabs=false,                      
    tabsize=2,                           
    frame=shadowbox,                         
    rulecolor=\color{gray},              
    framesep=2mm,                        
    xleftmargin=2mm,                     
    xrightmargin=2mm,                    
    aboveskip=2mm,                       
    belowskip=2mm,                       
    frameround=tttt                      
}
\usepackage[para]{footmisc}

\graphicspath{{figures/}}

\algnewcommand{\algcomment}[1]{\textit{// #1}}  
\newcommand{\figGap}[0]{\vspace{-\baselineskip}}

\definecolor{lightgray}{gray}{0.9}
\definecolor{lightgreen}{rgb}{0.8,1,0.8}
\definecolor{perfblue}{RGB}{64, 114, 175}
\definecolor{perfred}{RGB}{220, 60, 60}

\newcommand{\cellgreen}{\cellcolor{lightgreen}}

\newcommand{\best}{\cellcolor{lightgreen}}

\newcommand{\DAgger}{\textsc{DAgger}\xspace}
\newcommand{\method}{\texttt{LEAP}\xspace}
\newcommand{\react}{\texttt{ReAct}\xspace}
\newcommand{\gptfo}{\texttt{gpt-4o}\xspace}
\newcommand{\gptfomini}{\texttt{gpt-4o-mini}\xspace}
\newcommand{\gptf}{\texttt{gpt-4}\xspace}
\newcommand{\gptt}{\texttt{gpt-3.5}\xspace}

\newcommand{\rew}{r}

\theoremstyle{plain}
\newtheorem{theorem}{Theorem}[section]

\theoremstyle{definition}
\newtheorem{definition}[theorem]{Definition}

\theoremstyle{remark}

\title{Better than Your Teacher: LLM Agents \\ that learn from Privileged AI Feedback}

\author{
Sanjiban Choudhury$^{1,*}$, Paloma Sodhi$^{2,*}$ \\
$^{1}$Cornell University, NY, USA, 
$^{2}$ASAPP Research, NY, USA \\
\texttt{sanjibanc@cornell.edu, paloma.sodhi@gmail.com} \\
}

\makeatletter
\def\thanks#1{\protected@xdef\@thanks{\@thanks \protect\footnotetext{#1}}}
\makeatother

\thanks{$^{*}$ Equal contribution.}

\begin{document}
\doparttoc 
\faketableofcontents 
\part{} 

\maketitle

\begin{abstract}

While large language models (LLMs) show impressive decision-making abilities, current methods lack a mechanism for automatic self-improvement from errors during task execution. We propose \method, an iterative fine-tuning framework that continually improves LLM agents using feedback from AI expert teachers. Our key insight is to equip the expert teachers with a \emph{privileged state} \textemdash information available during training but hidden at test time. This allows even weak experts to provide precise guidance, significantly improving the student agent's performance without access to privileged information at test time.
We evaluate \method on diverse decision-making benchmarks, including text-based games (ALFWorld), web navigation (WebShop), and interactive coding (Intercode Bash). Our experiments show that \method (1) outperforms behavior cloning and \texttt{ReAct} baselines (2) enables weak student models (e.g., Llama3-8B) to exceed the performance of strong teacher models (GPT4-o), and (3) allows weak models to self-improve using privileged versions of themselves. We also provide a theoretical analysis showing that \method’s success hinges on balancing privileged information with the student’s realizability, which we empirically validate. Our code is available at \url{https://leap-llm.github.io}.

\end{abstract}

\vspace{-1em}
\section{Introduction}
\vspace{-0.5em}
Large language models (LLM) show impressive decision-making abilities across domains~\citep{yao2022react, jimenez2023swe, sodhi2024step}. However, they struggle to learn and recover from errors encountered at test time. Prompting alone is insufficient, as it requires manually specifying exceptions, which is difficult to scale and leads to intractably long context lengths.



We address the problem of fine-tuning LLM agents to improve their decision-making using online interaction data. Imitation learning offers a sample-efficient way to train such models, where a teacher corrects the student agent's actions. However, finding an expert capable of both demonstrating the task and correcting mistakes from any state the student visits is extremely challenging. The expert must not only know how to complete the task but also how to recover from the student's errors, which is often more complex than simply performing the original task. For example, in a task like "Heat the mug and put it in the cabinet," if the student places the wrong object in the microwave, the expert must first backtrack and remove the incorrect object before proceeding with the task.

\begin{figure*}[!t]
\centering
\includegraphics[width=0.9\textwidth, trim = 0cm 0.0cm 0cm 0cm]{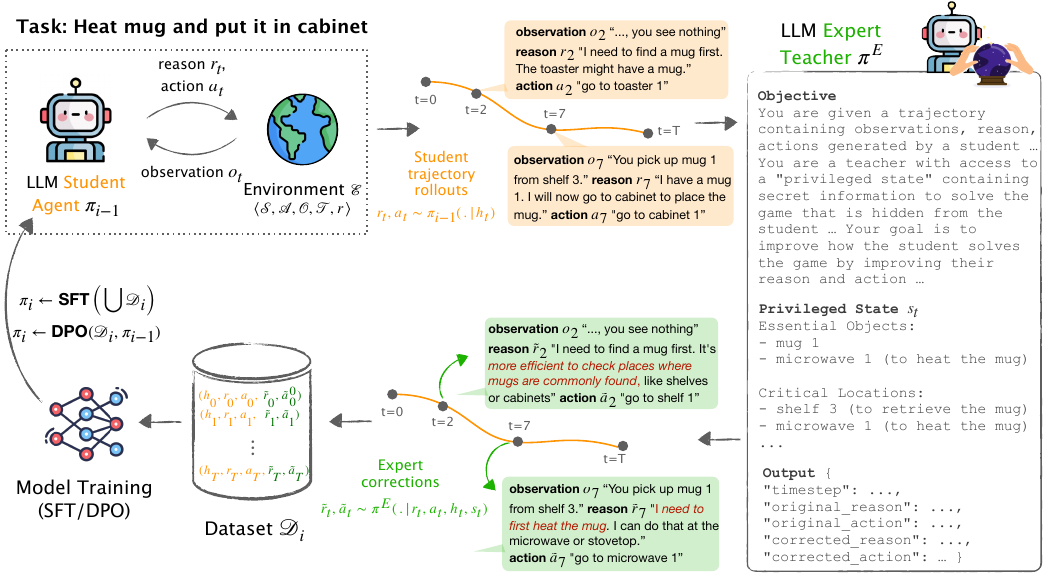}
\caption{\small \textbf{\method overview}. LLM student agent interacts with the environment, generating a reason-action trajectory (\textcolor{orange}{in orange}) based on its policy $\pi_{i-1}$. An expert teacher, with privileged state available only during training, evaluates and corrects the trajectory (\textcolor{green}{in green}). These corrections update the learner's policy to $\pi_{i}$ through SFT/DPO training. Updated policy $\pi_{i}$ is then rolled out at test time without access to privileged state. \figGap \vspace{-0.em}}
\label{fig:overall_approach}
\end{figure*}

\textbf{\emph{Our key insight is to equip the expert teacher with access to a privileged state}} \textemdash\ additional information unavailable to the student during test time, but crucial to improve the expert’s ability to offer corrective actions at train time. This may include hidden states of a partially observable environment or intermediate subgoals. For the task "Heat the mug and put it in the cabinet", the privileged state reveals the mug's location and necessary steps for heating it (Fig.~\ref{fig:overall_approach}). With this extra knowledge, the expert can first clear the microwave of any incorrect items and then teach the student effective search strategies, like checking common mug locations. Over time, this enables the student to not only learn how to perform tasks correctly but also how to recover from mistakes, ultimately allowing the student to outperform the expert, even without access to the privileged state at test time.

We propose \textbf{L}earning from \textbf{E}xperts with \textbf{A}ccess to \textbf{P}rivilege (\method), an iterative learning algorithm that fine-tunes LLM agents using privileged AI expert feedback. As shown in Fig.~\ref{fig:overall_approach}, \method begins with a student agent trained on demonstration data ($\pi_0$). During interactions, the student agent generates trajectories, and the privileged expert ($\pi^E$) identifies key time steps where corrections are needed. The expert provides improved reason and action labels based on the privileged state, which are then assembled into a dataset used to update the student model ($\pi_{i}\leftarrow\pi_{i-1}$). This process is repeated iteratively, progressively improving the student agent's performance.

Interestingly, we identify a key \emph{trade-off between leveraging privileged information and the student agent's realizability}. While providing unrestricted privileged feedback allows for highly accurate corrections, these are often too complex for the student to act on effectively. Conversely, limiting feedback to what the student can easily process makes it realizable but often suboptimal. Our theoretical and empirical analyses show that balancing these extremes is key, which we achieve by introducing a constrained privileged expert that provides optimal yet realizable feedback. 

Our key contributions are:
\begin{enumerate}[nosep, leftmargin=0.2in]
    \item A novel iterative learning framework \method that fine-tunes LLM agents using privileged expert feedback, balancing privileged information and realizability.
    \item Theoretical and empirical analysis of the optimal trade-off between privileged information and agent realizability.
    \item Experimental validation on diverse interactive decision-making benchmarks: AlfWorld~\citep{shridhar2020alfworld}, WebShop~\citep{yao2022webshop}, and InterCode~\citep{yang2024intercode}.
    \begin{enumerate}[nosep, leftmargin=0.3in]
        \item \method consistently outperforms behavior cloning agents trained on \gptfo or human demonstrations, showing significant improvements on ALFWorld ($65.7\%\rightarrow 91.8\%$), WebShop ($29.4\%\rightarrow 61.8\%$), and InterCode Bash ($60.3\%\rightarrow 71.8\%$).
        \item \method enables weaker models (e.g. \texttt{Llama3-8B}) to surpass stronger models (\gptfo), with improvements on ALFWorld ($91.8\%$ vs. $65.7\%$), WebShop ($61.8\%$ vs. $58.4\%$), and InterCode Bash ($71.8\%$ vs. $71.3\%$).
        \item Ablations that show LLM agents can self-improve using their privileged versions as teachers ($65.7\% \rightarrow 82.1\%$), explore the tradeoff between privileged information and realizability, and compare supervised finetuning with preference optimization.
    \end{enumerate}
\end{enumerate}

\vspace{-0.75em}
\section{Preliminaries}
\vspace{-0.5em}

\textbf{Problem Formulation: Decision-making LLMs.}
We frame decision-making as a Partially Observable Markov Decision Process (POMDP)~\citep{kaelbling1998planning} because real-world environments often contain hidden or incomplete information. The POMDP is defined by the tuple $\langle \mathcal{S}, \mathcal{A}, \mathcal{O}, \mathcal{T}, r \rangle$, where $\mathcal{S}$ represents the underlying state space of the environment, $\mathcal{A}$ the action space, $\mathcal{O}$ the observation space, $\mathcal{T}: \mathcal{S} \times \mathcal{A} \to \mathcal{S}$ the transition dynamics, and $r: \mathcal{S} \times \mathcal{A} \to \mathbb{R}$ the reward function. The true state $s_t \in \mathcal{S}$ is not fully observable; instead, the agent receives partial observations $o_t \in \mathcal{O}$ at each timestep. Hence it must rely on the history $h_t \in \mathcal{H}$ of observation-actions to choose actions.

LLM-based agents are policies $\pi \in \Pi$, which predict both a \emph{reason} $\rho_t$ and an action $a_t$ based on the agent's history $h_t$. The reason $\rho_t$ is expressed in natural language to justify action $a_t$. 
The history $h_t = {o_1, a_1, \dots, o_t}$ is the sequence of past observations and actions. The policy is expressed as: 
\begin{equation} 
(\rho_t, a_t) = \pi(h_t) 
\end{equation} 
While reinforcement learning (RL) for POMDPs is already challenging due to the large history space $\mathcal{H}$~\citep{papadimitriou1987complexity}, this is further exacerbated by the addition of reason space, which is infinite and unconstrained. This makes RL impractical to run for LLM agents.

 
\textbf{Interactive Imitation Learning with \DAgger.} Imitation learning (IL) offers a promising alternative to RL, by allowing the agent to mimic an expert teacher policy $\pi^E$ bypassing the need for complex exploration. The gold-standard for imitation learning are interactive methods like \DAgger~\citep{ross2011reduction} that query the teacher $\pi^E$ for feedback on states $h_t$ visited by the student. At each \DAgger iteration $i$, the student policy $\pi_i$ is rolled out to collect histories $h_t$, and the teacher provides corrective reasons and actions $(\rho^*_t, a^*_t)$ on those histories $h_t$. The new student $\pi_{i+1}$ then minimizes the cumulative loss $\ell(.)$ on all the data up until the current iteration $i$: 
\begin{equation} 
\pi_{i+1} = \min_{\pi} \quad \sum_{j=0}^{i}\mathbb{E}_{h_t \sim \pi_j, \rho^*_t, a^*_t \sim \pi^E(h_t)} \left[ \ell( (\rho^*_t, a^*_t), \pi(h_t)) \right], 
\end{equation} 
\DAgger guarantees that as iterations $N \rightarrow \infty$, it will find the best policy in policy class $\Pi$.

\vspace{-0.5em}
\section{Approach}
\vspace{-0.5em}
\label{sec:approach}
We present \textbf{L}earning from \textbf{E}xperts with \textbf{A}ccess to \textbf{P}rivilege (\method), a framework for iteratively training an LLM agent through online AI feedback from a privileged expert (Fig.~\ref{fig:overall_approach}). 
Central to our approach is the concept of a \emph{privileged expert teacher}, which leverages access to privileged state information to guide the student’s recovery from errors. Section~\ref{sec:approach:teacher} defines the privileged teacher, Section~\ref{sec:approach:algorithm} details \method, and Section~\ref{sec:approach:analysis} presents the theoretical analysis of \method. 
\vspace{-0.5em}
\subsection{Privileged Expert Teacher}
\label{sec:approach:teacher}
\vspace{-0.5em}
An expert teacher is a policy that can be queried with student history $h_t$ of observations and actions, along with the student's predicted reason action $(\rho_t, a_t)$ to generate a corrected reason action $(\tilde{\rho}_t, \tilde{a}_t) \sim \pi^E(\cdot | {r}_t, {a}_t, h_t)$. For POMDPs, computing such corrections is intractable and strictly more challenging than just solving the original task, often resulting in suboptimal feedback.

We define a \emph{privileged expert teacher} as a teacher that has access to privileged state information, $s_t$, which fully describes the environment at a given time. This access transforms the teacher's decision-making into a fully observable Markov Decision Process (MDP), making it significantly more tractable. We formally define the privileged expert teacher as: \begin{equation} (\tilde{\rho}_t, \tilde{a}_t) \sim \pi^E(\cdot | {\rho}_t, {a}_t, h_t, s_t) \end{equation}

The privileged state $s_t$ can either be explicitly provided or is implicit in privileged text information provided to the expert. For instance, in Alfworld~\citep{shridhar2020alfred}, it includes the exact location of objects and the goal necessary for the agent to navigate effectively. In WebShop~\citep{yao2022webshop}, it involves the specific name of the target product and its attributes, aiding precise search queries or recommendations. Similarly, in InterCode~\citep{yang2024intercode}, the privileged state consists of the ground truth code and any unit tests the code must pass, providing a clear benchmark for the agent to generate or correct code. Note that the privileged information is available only at train time.

A crucial requirement for the privileged expert teacher is that \emph{it must not reveal privileged information in its feedback to the learner.} The prompt instructions specify that the teacher should provide general reason and action that avoid the direct use of privileged information. Using such information directly might solve tasks quickly but makes the corrected actions $(\tilde{\rho}_t, \tilde{a}_t)$ unrealizable for the learner, who does not have access to the privileged state $s_t$, leading to a significant imitation gap and poor learner performance. In Section~\ref{sec:approach:analysis}, we characterize the trade-off between using privileged information and student realizability. We show in Appendix.~\ref{sec:appendix:analysis} that a constrained privileged expert that effectively utilizes privileged information while remaining realizable for the student has the optimal performance. We approximate such an expert by designing prompts for the expert to instruct it to use privileged information, without revealing it. See Appendix~\ref{sec:appendix:prompts} for prompts.

\vspace{-0.5em}
\subsection{The \method Algorithm}
\label{sec:approach:algorithm}

\begin{algorithm}[!t]
\caption{\method: Iterative Learning with Privileged Expert Teacher}
\begin{algorithmic}[1]
    \Statex \textbf{Input:} Privileged expert teacher $\pi^E$, Environment $\mathcal{E}$
    \Statex \textbf{Output:} Learned policy $\pi$
    \State Collect an initial set of demonstrations from expert teacher $\mathcal{D} \leftarrow \{ (h^*_t, \rho^*_t, a^*_t) \}$
    \State Train BC $\pi_0$ on $\mathcal{D}$ using supervised fine-tuning (SFT)
    \For{iteration $i = 1, 2, \dots, N$}
        \State Roll out policy $\pi_{i-1}$ in environment $\mathcal{E}$ to collect new data: $\mathcal{D}_i \leftarrow \{(h_t, \rho_t, a_t)\}$
        \State Compute privileged state $s_t$ for every datapoint in $\mathcal{D}_i$
        \State Invoke privileged expert to generate corrected reason and action: $\tilde{\rho}_t, \tilde{a}_t \sim \pi^E ( . | {\rho}_t, {a}_t, h_t, s_t)$
        \State Augment dataset with corrected reason and action: $\mathcal{D}_i \leftarrow \{(h_t, \rho_t, a_t, \tilde{\rho}_t, \tilde{a}_t)\}$
        \State Train policy $\pi_i$ using a no-regret update: $\pi_i \gets \text{SFT}\left(\bigcup \mathcal{D}_i\right)$ or $\pi_{i} \gets \text{DPO}(\mathcal{D}_i, \pi_{i-1})$
    \EndFor
    \Statex
    \Return Best $\pi \in \{\pi_1, \dots, \pi_N\}$ on validation dataset
\end{algorithmic}
\end{algorithm}

The \method algorithm iteratively trains the student policy $\pi$ through online interactions with the environment $\mathcal{E}$ and corrections from a privileged expert teacher $\pi^E$. The process begins by collecting an initial set of demonstrations of history, reason, and action from the teacher $\mathcal{D} \gets \{h^*_t, \rho^*_t, a^*_t\}$, and training a BC policy $\pi_0$ on this data using supervised fine-tuning (SFT). 

At each iteration, the current learner policy $\pi_i$ is rolled out in the environment to generate new data $\mathcal{D}_i \leftarrow {(h_t, \rho_t, a_t)}$. For each timestep in $\mathcal{D}_i$, \method computes the privileged state $s_t$ by leveraging information available only during training. The privileged teacher is then invoked on trajectories where the agent fails, generating corrected reasoning and actions, $\tilde{\rho}_t, \tilde{a}_t \sim \pi^E(. | \rho_t, a_t, h_t, s_t)$. These corrections are used to augment the dataset $\mathcal{D}_i$. The policy $\pi_i$ is subsequently updated using this augmented dataset through a no-regret learning method. This iterative process continues, refining the policy over multiple iterations until the best-performing policy is selected based on validation.

The update step in \method should be designed to ensure no-regret learning, a property that guarantees the policy's cumulative performance will converge to that of the best possible policy in hindsight. The most common update is to aggregate data $\bigcup \mathcal{D}_i$, and train $\pi_i$ using supervised fine-tuning (SFT) on the aggregated dataset, as done in \DAgger~\citep{ross2011reduction}. This is equivalent to Follow the Regularized Leader (FTRL), a no-regret update step. An alternate update is to treat the problem as a preference optimization problem where the corrected reason action $(\tilde{\rho}_t, \tilde{a}_t)$ is preferred over the student reason action $(\rho_t, a_t)$. Preference optimizers like Direct Policy Optimization (DPO)~\citep{rafailov2024direct} or Kahneman-Tversky Optimization (KTO)~\citep{ethayarajh2024kto} optimize a preference loss while regularizing to the base policy, in this case the previous policy $\pi_{i-1}$. This is equivalent to an Online Mirror Descent (OMD) update, which is also a no-regret update. We use SFT by default but compare it against preference optimization as an ablation in the results.
\vspace{-0.5em}
\subsection{Analysis}
\vspace{-0.5em}
\label{sec:approach:analysis}
We briefly touch on the theoretical guarantees of \method, and refer the reader to Appendix.~\ref{sec:appendix:analysis} for a full exposition. We define the realizability of an expert policy and show how it affects \method performance.
\begin{definition}[Average Realizability Gap] The average realizability gap $epsilon (\pi^E, T)$ between the privileged expert $\pi^E$ and the best policy $\pi^\star$ over time horizon $T$ is:
\begin{equation}
\epsilon (\pi^E, T)  := \sup_{\pi} \frac{1}{T} \sum_{t=1}^{T} \mathbb{E}_{s_t, h_t \sim d^\pi_t} || \pi^E(a_t|s_t) - \pi^{\star}(a_t | h_t)) ||_1
\end{equation}
where $||.||_1$ is the L1 distance, $d^\pi_t$ is the induced distribution over state and history of \emph{any policy $\pi$}.
\end{definition}
\begin{theorem}[\method with Privileged Expert] Running $N$ iterations of \method with privileged expert $\pi^E$ yields at least one policy $\pi$ such that the time-average performance $\frac{1}{T} J(\pi)$ is bounded as:
\begin{equation}
    \frac{1}{T} J(\pi) \geq \frac{1}{T} J(\pi^E)  - H(\pi^E) \left( \epsilon (\pi^E, T) + \gamma(N) \right)
\end{equation}
where $\pi^E$ is the privileged expert, $H(\pi^E)$ is the recoverability coefficient of $\pi^E$, $\epsilon (\pi^E, T)$ is the average realizability gap, and $\gamma(N)$ is the average regret of the \DAgger update. 
\end{theorem}
The equation shows a trade-off between performance of the privileged expert (first term) and the realizability of the privileged expert (second term), which we show how to optimize in 
Appendix.~\ref{sec:appendix:analysis}.

\vspace{-0.5em}
\section{Experiments}
\label{sec:experiments}

\vspace{-0.5em}
\subsection{Overview of Results}
\vspace{-0.5em}

We evaluate \method across $3$ decision-making domains: ALFWorld, a text-based game, WebShop, a web navigation environment, and Intercode, an interactive coding environment. Below is a summary:

\begin{enumerate}[nosep, left=0.2mm]
    \item \textbf{\method improves upon base behavior cloning (BC) policy:} \method significantly improves the base BC policy $\pi_0$ over successive fine-tuning iterations, showing gains on AlfWorld ($65.7\%\rightarrow 91.8\%$), WebShop ($29.4\%\rightarrow 61.8\%$), and InterCode ($60.3\%\rightarrow 71.8\%$). See Tables~\ref{tab:alfworld_eval_in_and_out_of_distribution},~\ref{tab:intercode_bash_eval}, Fig.~\ref{fig:webshop_eval_table_plots}.
    \item \textbf{\method enables weaker student models to surpass stronger teacher models:} Over successive iterations, student models (e.g., Llama3-8B) outperform their stronger teachers (e.g., GPT-4o), despite starting with lower or comparable performance. For example, on ALFWorld, the student model ($\pi_3: 91.8\%$) surpasses the teacher ($\pi^E: 65.7\%$), with similar trends on WebShop ($\pi_2: 61.8\%$ vs. $\pi^E: 58.4\%$) and InterCode ($\pi_3: 71.8\%$ vs. $\pi^E: 71.3\%$). See Tables~\ref{tab:alfworld_eval_in_and_out_of_distribution},~\ref{tab:intercode_bash_eval}, Fig.~\ref{fig:webshop_eval_table_plots}.
    \item \textbf{\method requires a balance between using privileged information and generating realizable corrections:} Experts using minimal privileged information ($\pi^E_1: 73.1\%$) generate more realizable actions, while those with excessive privileged feedback ($\pi^E_5: 5.22\%$) produce unrealizable corrections. $\pi^E_3$ strikes a perfect balance achieving optimal performance ($\pi^E_3: 91\%$). See Fig.~\ref{fig:ablation_privileged_info_combined}.
    \item \textbf{\method enables LLM agents to self-improve by using their own privileged versions as experts:} LLM agents can self-improve by using their privileged versions as teachers. For example, in ALFWorld, Llama3-8B bootstraps itself through iterations, improving from $65.7\%$ to $82.1\%$ and finally to $91.8\%$. See Fig.~\ref{fig:self_correction_plot_alfworld}.    
    \item \textbf{\method can be trained using both SFT or Preference Optimization:} Both SFT and preference optimization improve over the base BC policy ($\pi_0$), with SFT delivering stronger gains ($68.6\%\rightarrow 96.4\%$) compared to DPO ($68.6\%\rightarrow 74.3\%$) and KTO ($68.6\%\rightarrow 87.1\%$). See Table~\ref{tab:sft_vs_preference}.
\end{enumerate}

\vspace{-0.5em}
\subsection{Domain 1: Text-based games}
\vspace{-0.5em}
\textbf{Setup.} ALFWorld~\citep{shridhar2020alfworld} is a text-based game where each task has a high-level objective, e.g. ``heat mug and put it in cabinet'' which the agent must complete by navigating and interacting with a virtual household through text commands, e.g. go to shelf 1, pick up mug 2, etc. To solve it, the agent must plan subgoals, track progress, and search for objects efficiently (e.g. mugs are more likely to be on shelves or cabinets). Each task has $30$ timesteps. There are $6$ categories of tasks, training dataset of $3257$ games and two evaluation datasets: $139$ in-distribution games and $134$ out-of-distribution games. We compare against a prior work BUTLER~\citep{shridhar2020alfworld} and \texttt{ReAct}~\citep{yao2022react}; the original \react paper used text-davinci-002 with task-dependent in-context examples. We add \react with stronger models such as \gptfo\footnote{https://platform.openai.com/docs/models}, claude\footnote{https://docs.anthropic.com/en/docs/about-claude/models}, and gemini\footnote{https://ai.google.dev/gemini-api/docs/models/gemini}, and a general instruction prompt. For \method, we LoRA~\citep{hu2021lora} fine-tune Llama3-8B~\citep{dubey2024llama} model using \gptfo as the privileged expert teacher for $4$ iterations. The privileged information for each game contains the essential objects to solve the task, object locations, and the optimal actions.  See Appendix.~\ref{sec:appendix:setup:alfworld},~\ref{sec:appendix:prompts:alfworld}  for full details and prompts.

\begin{table}[!t]
\centering
\begin{minipage}{\textwidth}
\centering
\resizebox{0.9\textwidth}{!}{
\begin{tabular}{l|cc|c c c c c c}
\toprule
\multirow{2}{*}{Method} & \multicolumn{2}{c|}{\textbf{All tasks}} & \multicolumn{1}{c}{\textsc{Pick} tasks} & \multicolumn{1}{c}{\textsc{Clean} tasks} & \multicolumn{1}{c}{\textsc{Heat} tasks} & \multicolumn{1}{c}{\textsc{Cool} tasks} & \multicolumn{1}{c}{\textsc{Look} tasks} & \multicolumn{1}{c}{\textsc{Pick 2} tasks} \\
\cmidrule(lr){2-9}
& \texttt{\%suc}$\uparrow$ & \texttt{\#act}$\downarrow$ & \texttt{\%suc}$\uparrow$ & \texttt{\%suc}$\uparrow$  & \texttt{\%suc}$\uparrow$  & \texttt{\%suc}$\uparrow$  & \texttt{\%suc}$\uparrow$  & \texttt{\%suc}$\uparrow$ \\
\midrule
\texttt{BUTLER} \textcolor{perfred}{[1]}  & 35.0 & - & 50.0 & 74.0 & 83.0 & 91.0 & 39.0 & 65.0 \\
\texttt{ReAct} few-shot \textcolor{perfred}{[2]} & 57.0 & - & 65.0 & 39.0 & 83.0 & 76.0 & 55.0 & 24.0 \\
\cmidrule(lr){1-9}
\texttt{ReAct} \gptfo & 65.7 & 20.2 & 91.7  & 35.5  & 56.5  & 52.4  & 100.0  & 76.5 \\
\texttt{ReAct} \gptfomini & 29.9 & 25.5 & 33.3  & 25.8  & 17.4  & 14.3  & 66.7  & 29.4 \\
\texttt{ReAct} \texttt{claude-3.5-sonnet} & 76.1 & 19.0 & \cellgreen 95.8  & 61.3  & 60.9  & 81.0  & 88.9  & 76.5 \\
\texttt{ReAct} \texttt{claude-3.5-haiku} & 16.4 & 27.2 & 33.3  & 9.7  & 8.7  & 9.5  & 38.9  & 0.0 \\
\texttt{ReAct} \texttt{gemini-1.5-flash} & 19.4 & 26.3 & 41.7  & 12.9  & 13.0  & 19.0  & 16.7  & 11.8 \\
\midrule
\method \texttt{Llama3-8B} & 0.7 & 29.8 & 0.0  & 0.0  & 0.0  & 4.8  & 0.0  & 0.0 \\
\method \texttt{Llama3-8B} $\pi_0$ & 65.7 & 18.6 & 66.7  & 74.2  & 73.9  & 66.7  & 66.0  & 35.3 \\
\method \texttt{Llama3-8B} $\pi_1$ & 91.0 &  11.9 & 83.3  & 90.3  & \cellgreen 91.3  & \cellgreen 95.2  & \cellgreen 94.4  & \cellgreen 94.1 \\
\method \texttt{Llama3-8B} $\pi_2$ & 83.6 & 13.1 &  87.5  & 90.3  & 73.9  & \cellgreen 95.2  & 66.7  & 82.4 \\
\method \texttt{Llama3-8B} $\pi_3$ & \cellgreen 91.8 & \cellgreen 11.3 &  87.5 & \cellgreen 93.5  & \cellgreen 91.3  & 90.5  & \cellgreen 94.4  & \cellgreen 94.1 \\
\bottomrule
\end{tabular}
}
\end{minipage}
\vspace{0.1cm}
\begin{minipage}{\textwidth}
\centering
\resizebox{0.9\textwidth}{!}{
\begin{tabular}{l|cc|c c c c c c}
\toprule
\multirow{2}{*}{Method} & \multicolumn{2}{c|}{\textbf{All tasks}} & \multicolumn{1}{c}{\textsc{Pick} tasks} & \multicolumn{1}{c}{\textsc{Clean} tasks} & \multicolumn{1}{c}{\textsc{Heat} tasks} & \multicolumn{1}{c}{\textsc{Cool} tasks} & \multicolumn{1}{c}{\textsc{Look} tasks} & \multicolumn{1}{c}{\textsc{Pick 2} tasks} \\
\cmidrule(lr){2-9}
& \texttt{\%suc}$\uparrow$ & \texttt{\#act}$\downarrow$ & \texttt{\%suc}$\uparrow$ & \texttt{\%suc}$\uparrow$  & \texttt{\%suc}$\uparrow$  & \texttt{\%suc}$\uparrow$  & \texttt{\%suc}$\uparrow$  & \texttt{\%suc}$\uparrow$ \\
\midrule
\texttt{BUTLER} \textcolor{perfred}{[1]}  & 40.0 & - & 69.0 & 67.0 & 88.0 & 76.0 & 69.0 & 54.0 \\
\texttt{ReAct} \gptfo & 54.3 & 22.9 & 91.4 & 33.3 & 31.2 & 12.0 & 84.6 & 66.7 \\
\texttt{ReAct} \gptfomini \phantom{heres} & 40.0 & 26.1 & 11.1 & 12.5 & 4.0 & 53.8 & 20.8 &  20.8 \\
\midrule
\method \texttt{Llama3-8B} & 2.9 & 29.5 & 8.6 & 0.0 & 0.0 & 0.0 & 7.7 & 0.0 \\
\method \texttt{Llama3-8B} $\pi_0$ & 68.6 & 16.9 & 88.6 & 51.9 & 56.2 & 80.0 & 76.9 & 50.0 \\
\method \texttt{Llama3-8B} $\pi_1$ & \best 96.4 &  10.5 & \best 100.0 & \best 100.0 & \best 100.0 & \best 92.0 & 92.3 & \best 91.7 \\
\method \texttt{Llama3-8B} $\pi_2$ & 90.0 & 11.4 & 94.3 & 92.6 &  87.5 & 88.0 &  92.3 & 83.3 \\
\method \texttt{Llama3-8B} $\pi_3$ & 94.2 & \cellgreen 10.3 & \best 100.0 & 96.3 & \best 100.0 & 87.5 & \best 100.0 & 83.3 \\
\bottomrule
\end{tabular}
}
\caption{ \textbf{AlfWorld Evaluation.} \textbf{(top)} $136$ out-of-distribution games and \textbf{(bottom)} 140 in-distribution  (max $30$ actions). Baseline comparisons include \textcolor{perfred}{[1]} \texttt{BUTLER}~\citep{shridhar2020alfred}, \textcolor{perfred}{[2]} \textsc{ReAct} few-shot~\citep{yao2022react}, and our own \textsc{ReAct} instruction prompt with different models. \method with an 8B model across iterations ($\pi_1, \pi_2, \pi_3$) outperforms the stronger teacher \react \gptfo. \figGap \vspace{-0.4em} }
\label{tab:alfworld_eval_in_and_out_of_distribution}
\end{minipage}
\end{table}

\begin{figure*}[!t]
\centering
\includegraphics[width=0.9\textwidth, trim = 0cm 0.0cm 0cm 0cm]{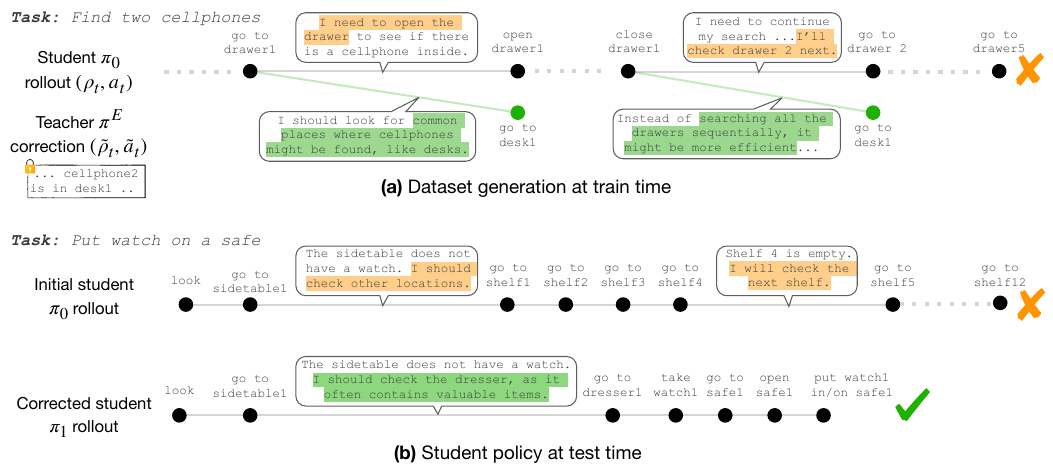}
\caption{\textbf{ALFWorld Training and Testing for \method.} \textbf{(a)} \textbf{Training:} Student policy $\pi_0$ rolled out on training task to generate reason and actions, e.g. it fails to cell phone because it inefficiently searches all drawers. Expert teacher $\pi^E$ uses privileged information (cellphone in desk1) to generate general corrected reason actions that don't reveal privileged information (cellphones commonly found in desks, more efficient to search desks than drawers).
 \textbf{(b)} \textbf{Testing:} $\pi_0$ fails to find a watch as it inefficently explores shelves one by one. $\pi_1$ learns a more efficient exploration policy, prioritizing areas like sidetables and dressers, solving the task quickly.\figGap \vspace{-2.5em}}
\label{fig:alfworld_qualitative}
\end{figure*}

\textbf{Results.}
Table~\ref{tab:alfworld_eval_in_and_out_of_distribution} shows that \method significantly outperforms all baselines, with the best policy achieving $91.8\%$ success rate on out-of-distribution tasks. Iteration $1$ of \method has the biggest performance gain $(65.7\% \rightarrow 91.0\%)$, \textbf{leading to a  student policy $\pi_1$ that surpasses the teacher \gptfo} with higher success rate $(91.8\% > 65.7\%)$ and lower actions $(11.3 < 20.2)$. Note that performance improvements are not monotonic over iterations of \method, $(65.7\% \rightarrow 91.0\% \rightarrow 83.6\% \rightarrow 91.8\%)$, however, all policies are better than the BC policy $\pi_0$. The regression from $\pi_1$ to $\pi_2$ are mainly from \textsc{Heat} and \textsc{Look} tasks, which gets corrected immediately in the next iteration $\pi_3$. Finally. we note that \method policies generalize well from in-distribution to out-of-distribution tasks with only a slight dip in performance for the final policy $\pi_3$ $(94.2\% \rightarrow 91.8\%)$. 

Fig.~\ref{fig:alfworld_qualitative} (a) shows an example of \method training and testing. A typical failure mode of $\pi_0$ is that it searches for items inefficiently, e.g. looking at shelves or drawers or cabinets one by one until time runs out. The expert teacher leverages the privileged state of where objects are to show the student how to search efficiently without revealing the object location, e.g. it's more efficient to search desks where cellphones are likely to be than drawers sequentially one after another. Training on this correction data, $\pi_1$ \emph{generalizes this reasoning to new test-time tasks}, e.g. putting a watch in a safe. Fig.~\ref{fig:alfworld_qualitative} (b) shows that while $\pi_0$ inefficiently searches shelves, $\pi_1$ searches likely locations for the watch, first the sidetable and then the dresser to solve the task. The main performance gains of \method over \gptfo come from such efficient exploration. \gptfo searches likely locations, however, its prior is from internet data and isn't necessarily aligned with the statistics of the ALFWorld environment. In contrast, \method leverages privileged feedback to learn this. See Appendix~\ref{sec:appendix:experiment_details} for more results. 

\vspace{-0.5em}
\subsection{Domain 2: Web agents}
\label{sec:results:webshop}
\vspace{-0.5em}

\begin{figure*}[!t]
\centering
\includegraphics[width=0.87\textwidth, trim = 0cm 0.0cm 0cm 0cm]{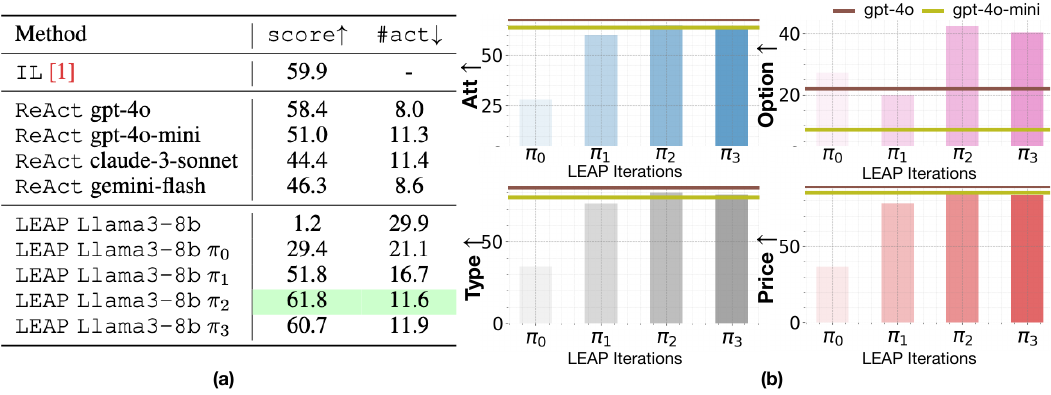}
\caption{\textbf{WebShop Evaluation.} \textbf{(a)} Overall \texttt{score}$\uparrow$ and \texttt{\#act}$\downarrow$ on $500$ test tasks (max $30$ actions). \textbf{(b)} Performance of \method over iterations on $4$ different score components. 
Baseline comparisons include \textcolor{perfred}{[1]} IL~\citep{yao2022webshop} and our \react instruction prompt with different models. \method with an 8B model across iterations ($\pi_2, \pi_3$) outperforms the stronger teacher \texttt{ReAct} \gptfo. \figGap}
\label{fig:webshop_eval_table_plots}
\end{figure*}

\textbf{Setup.} WebShop~\citep{yao2022webshop} is an online shopping environment where each task is a shopping request from a user (``I need a gluten-free bacon, 4 oz pack of 2, priced under \$50'') that the agent has to solve through web interactions (search[``gluten-free bacon''] or click options). Success is determined by an overall score with $4$ components: whether attributes match (\textsc{Att}), whether options match (\textsc{Opt}), whether product types match (\textsc{Type}) and whether price matches (\textsc{Price}). Each task has $30$ timesteps. The training dataset has $12086$ tasks, the test dataset has $500$ tasks. We compare against \texttt{ReAct}~\citep{yao2022react} with different base models like GPT-4o and GPT-4o-mini. We also compare against IL baseline in WebShop~\citep{yao2022webshop}, which trains two models, one for search (from a list of keywords) and one for clicking, thus simplifying the problem. 
We LoRA fine-tune Llama3-8B model using \gptfo as the privileged expert teacher for 4 iterations. The privileged information contains product attributes, product option, price constraint, and an example product that would satisfy all criteria. See Appendix.~\ref{sec:appendix:setup:webshop},~\ref{sec:appendix:prompts:webshop}  for full details and prompts.

\begin{figure*}[!t]
\centering
\includegraphics[width=0.9\textwidth]{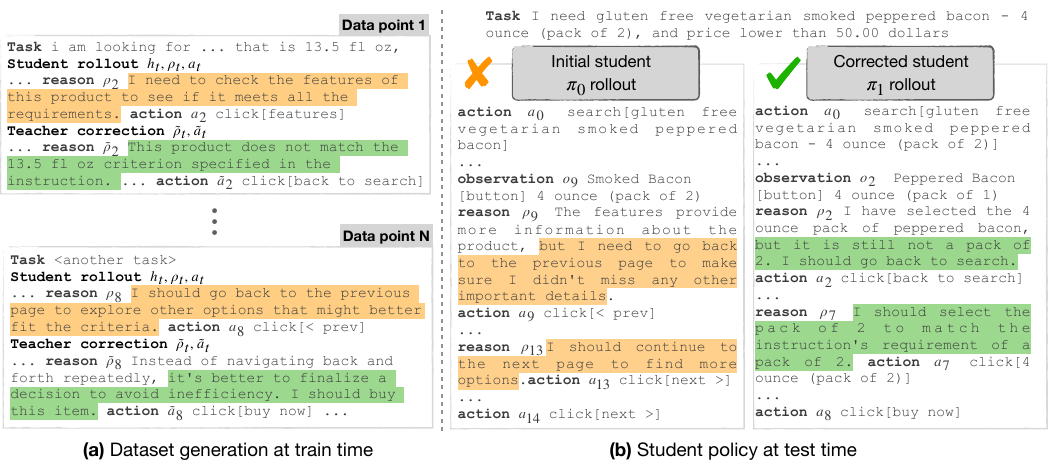}x
\caption{\small \textbf{WebShop Training and Testing for \method.} \textbf{(a) Training:} Teacher policy generates corrections on student rollout to backtrack when the product does not fit criteria or to commit to a product when it does. 
\textbf{(b) Evaluation:} $\pi_0$ fails to solve the task since it continues to search page after page even after discovering a good product. $\pi_1$ learns when to backtrack and when to commit to a product to solve the task in time. \figGap \vspace{-1.3em}
}
\label{fig:webshop_qualitative}
\end{figure*}

\textbf{Results.}
Fig.~\ref{fig:webshop_eval_table_plots} (a) shows that \method outperforms all baselines, with the best policy ($\pi_2$) achieving $61.4$ score. Iteration $1$ has the biggest performance gain $(29.4 \rightarrow 51.8)$ followed by iteration 2 $(51.8 \rightarrow 61.8)$. This \textbf{leads to a student policy $\pi_2$ that surpasses the teacher GPT-4o} with a higher score $(61.8 > 58.4)$. Similar to ALFWorld, while performance improvements are not monotonic, both $\pi_2, \pi_3$ remain above the teacher \gptfo performance. Fig.~\ref{fig:webshop_eval_table_plots} (b) shows how \method improves on all $4$ components of the score. The biggest improvements between \method and \gptfo is in \textsc{Opt} component $(42.4 > 22.1)$, indicating \gptfo fails to select a product with the right option often.

Fig.~\ref{fig:webshop_qualitative} (a) shows an example \method correction of $\pi_0$ and $\pi_1$ from the training data. The teacher leverages the privileged state such as the attributes or options of the product to teach the student to recognize when an item does not fit the description and the agent should backtrack. It also teaches the student when to commit to a product if it satisfies all criteria rather than being non-committal and searching for options. Training on this correction data yields a policy $\pi_2$ that balances correctly backtracking vs committing to a product. Fig.~\ref{fig:webshop_qualitative} (b) shows that $\pi_0$ fails to commit to a product needlessly cycling through pages till time runs out. In contrast, $\pi_2$ correctly backtracks and then selects a second item that solves the task. 

\vspace{-0.5em}
\subsection{Domain 3: Interactive Coding}
\vspace{-0.5em}

\textbf{Setup.} Intercode Bash~\citep{yang2024intercode} is an interactive coding environment, where the agent receives a task that involves operations on a filesystem, e.g. ``Find all text files and write their names to a single file'', and must interact with Bash to solve it and submit a final command.
\begin{wraptable}[15]{r}{0.4\textwidth}
\centering
\resizebox{0.4\textwidth}{!}{
\begin{tabular}{l|cc}
\toprule
Method & \texttt{\%succ}$\uparrow$ & \texttt{\#act}$\downarrow$ \\
\midrule
\texttt{gpt-4} \textcolor{perfred}{[1]} & $48.5^\star$ & - \\
\texttt{gpt-3.5} \textcolor{perfred}{[1]} & $46.5^\star$ & -\\
CodeLlama-34B-INST \textcolor{perfred}{[1]} & $36.0^\star$ & -\\
\midrule
\texttt{ReAct} \texttt{gpt-4o} & 71.3 & \cellgreen 3.8 \\
\texttt{ReAct} \texttt{gpt-4o-mini} & 43.4 & 6.0\\
\midrule
\method \texttt{Llama-3.1-8b} $\pi_0$ & 41.4 & 6.6 \\
\method \texttt{Llama-3.1-8b} $\pi_1$ & 49.9 & 5.5 \\
\midrule
\method \texttt{Llama-3.1-70b} $\pi_0$ &  60.3 &  4.8 \\
\method \texttt{Llama-3.1-70b} $\pi_1$ & \cellgreen 71.8 & 5.1 \\
\bottomrule
\end{tabular}
}
\caption{\textbf{Intercode Bash Evaluation.} Overall success rate and actions on test data (max $10$ actions). Baselines from \textcolor{perfred}{[1]} \citep{yang2024intercode} evaluated on $^\star$(train + test) data. \figGap}
\label{tab:intercode_bash_eval}
\end{wraptable}
Intercode uses NL2Bash~\citep{lin2018nl2bash} to create $4$ datasets. We use the first $2$ for training, and the next $2$ for testing. We limit time horizon to $10$. We compare against \react~\citep{yao2022react} with \gptfo and \gptfomini. 
We also compare against the top 3 entries on the Intercode leaderboard, which are the \texttt{TryAgain} baselines from \citep{yang2024intercode} with \gptf, \gptt and CodeLlama-34B-INST~\citep{roziere2023code} respectively. We LoRA fine-tune both Llama-3.1-8B model and Llama-3.1-70B model, each for $2$ iterations,  using \gptfo as the privileged expert teacher. The privileged information is the ground truth Bash command.

\textbf{Results.} Table~\ref{tab:intercode_bash_eval} shows that \method trained with $70$B policy outperforms all baselines, with policy $\pi_1$ achieving $71.8\%$ success rate to match the \gptfo expert teacher $(71.8\% > 71.3 \%)$. We find that on this coding benchmark, the \gptfo expert is a very strong baseline. \method trained with $8$B models improve over the BC policy, but falls short of the \gptfo expert teacher. These results show that if the expert teacher is already a strong baseline, even after leveraging privileged information, it might be hard for the student to outperform the teacher. 

\vspace{-0.5em}
\subsection{What is the trade-off between Privileged Information and Realizability?}
\label{sec:results:tradeoff}
\vspace{-0.5em}

\begin{figure*}[!t]
\centering
\includegraphics[width=0.9\textwidth, trim = 0cm 0.0cm 0cm 0cm]{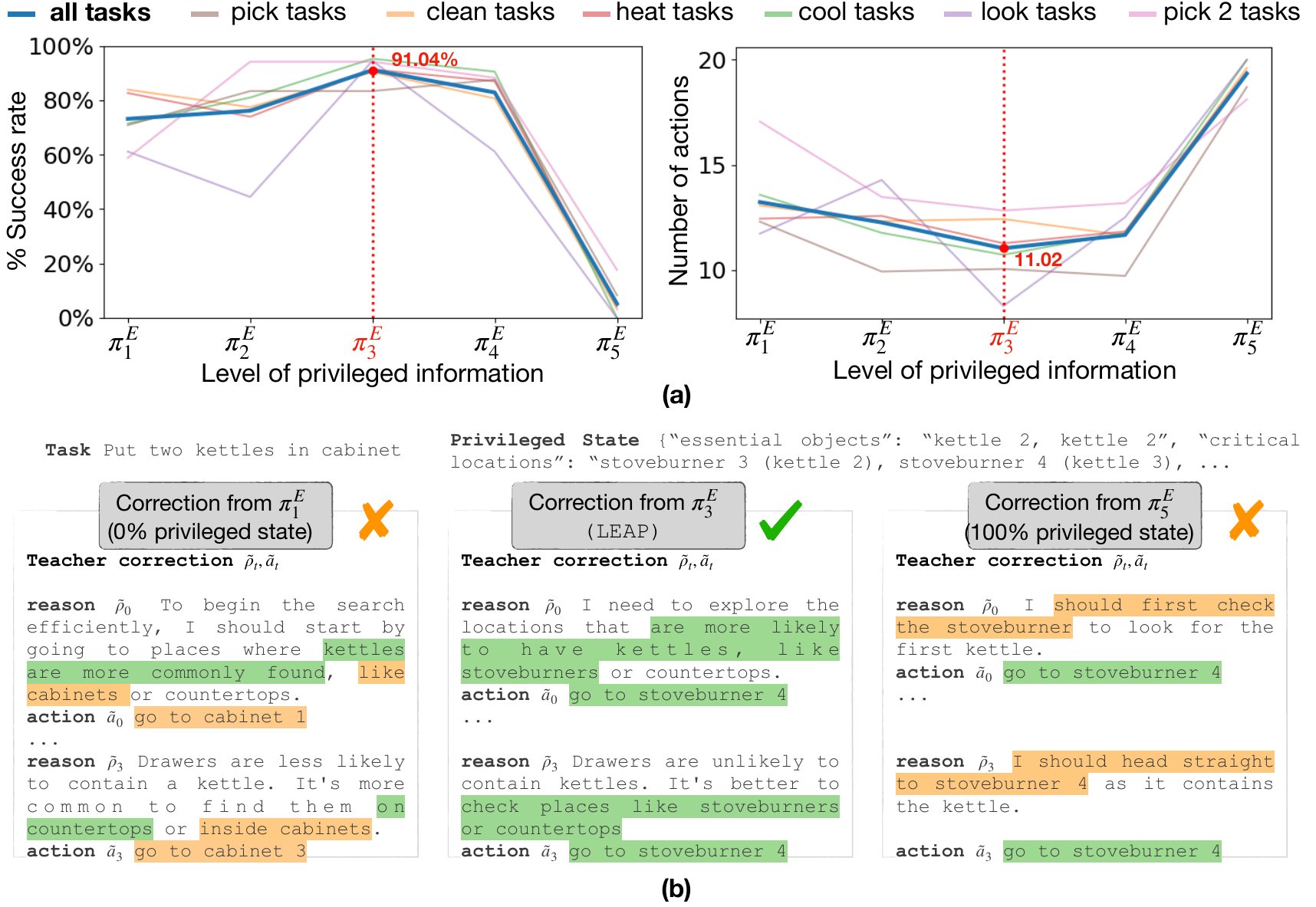}
\caption{\textbf{Privileged Information vs Realizability.} \textbf{(a)} Performance of $5$ policies trained with experts with varying levels of privileged information on ALFWorld, peaking for expert $\pi^E_3$. 
\textbf{(b)} Examples of corrections from experts $\pi^E_1, \pi^E_3, \pi^E_5$. $\pi^E_1$ generates realizable reason action but predicts wrong action. $\pi^E_5$ predicts correction action, but produces unrealizable reason action that contains privileged information. $\pi^E_3$ strikes a perfect balance. 
\figGap}
\label{fig:ablation_privileged_info_combined}
\figGap
\end{figure*}

We study the tradeoff between how much privileged state the expert uses vs the realizability of their corrections on ALFworld. We create $5$ different privileged experts $\pi_E^1, \dots, \pi_E^5$ by varing the prompt to instruct them to use increasing amounts of privileged state, e.g. for $\pi_E^2$ it says ``Use this information sparingly for your correction,'' but for $\pi_E^4$ it says ``feel free to include information from the privileged state''. We run 1 iteration of \method to generate updated $\pi^1, \dots, \pi^5$. Fig.~\ref{fig:ablation_privileged_info_combined} (a) shows that success rate initially rises from $\pi^1$ to $\pi^3$, but then falls off sharply till $\pi^5$. Num actions has a similar trend. This validates the hypothesis that there is indeed an optimal tradeoff. Interestingly, $\pi_5$ is much worse than $\pi_1$, showing that it's more important to be realizable than to provide optimal yet unrealizable corrections. Fig.~\ref{fig:ablation_privileged_info_combined} (b) shows the corrections from $\pi^E_1, \pi^E_3, \pi^E_5$. $\pi^E_1$ generates reasoning that is perfectly realizable to the student, but since it has no privileged state the actions are not optimal. In contrast, $\pi^E_5$ predicts the correct action, but blatantly reveals the privileged information in the reason, being unrealizable to the student. $\pi^E_3$ strikes a perfect balance where it offers the correct action, but offers a reason that is very much realizable, i.e. kettles are likely to be in stoveburners.

\vspace{-0.5em}
\subsection{Can \method be used to self-correct a student?}
\vspace{-0.5em}
\begin{figure*}[!t]
\centering
\includegraphics[width=0.9\textwidth, trim = 0cm 0.0cm 0cm 0cm]{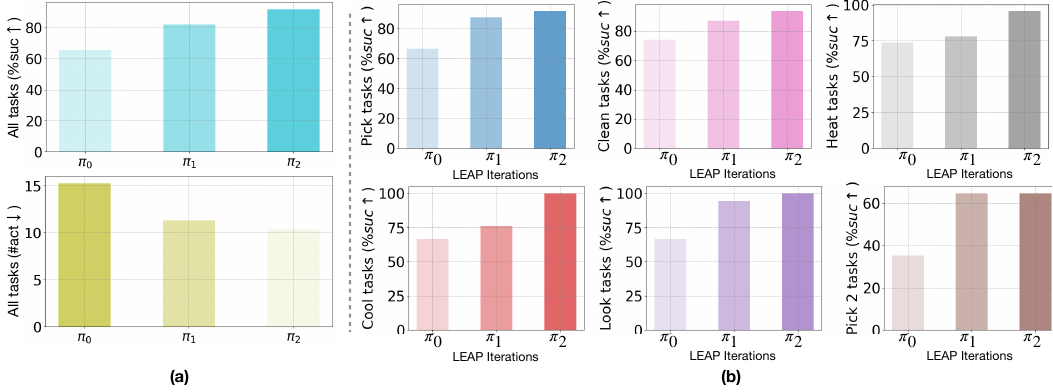}
 \caption{ \textbf{Self Correction.} \method self-improves a Llama3-8B model that is used as both student and teacher. \textbf{(a)} Overall success rate and num actions. \textbf{(b)} Category-wise breakdown of success rate. \figGap \vspace{-1.1em}}
\label{fig:self_correction_plot_alfworld}
\end{figure*}

We test the hypothesis that \method should enable a model to self-improve by using its privileged version as the expert. On ALFWorld, we fine-tune Llama3-8B initializing with the BC policy $\pi_0$ and running $2$ iterations of self-improvement. Fig.~\ref{fig:self_correction_plot_alfworld} shows that \method is able to significantly self-improve policies over iterations $(65.7\% \rightarrow 82.1\% \rightarrow 91.8\%)$, with the final policy \textbf{matching the best policy when using GPT-4o as a teacher}. We see uniform improvements across all categories, with some being improved more from $\pi_0 \rightarrow \pi_1$ and others from $\pi_1 \rightarrow \pi_2$. We conclude that for some environments, like ALFworld, the privileged state does the heavy lifting, i.e. performance improvements happen at a similar rate to using a strong teacher vs using the model itself as a teacher. We note that this need not hold for all environments, e.g. for some a strong teacher is required to extract the optimal reason and action from the privileged state. Additionally, if the base policy $\pi_0$ is not strong, there maybe not lift off for the model and rate of improvement would be limited. 

\vspace{-0.5em}
\subsection{How does SFT compare to Preference Optimization?}
\vspace{-0.5em}

\begin{wraptable}{r}{0.4\textwidth}
\centering
\resizebox{0.4\textwidth}{!}{
\begin{tabular}{l|l|cc}
\toprule
 & Method & \texttt{\%succ}$\uparrow$ & \texttt{\#act}$\downarrow$ \\
\midrule
\multirow{4}{*}{\rotatebox[origin=c]{90}{in-dist.}} 
& \method $\pi_0$ & 68.6 & 16.9 \\
& \method \texttt{DPO} ($\beta:0.1$) $\pi_1$ & 74.3 & 15.7 \\
& \method \texttt{DPO} ($\beta:0.01$) $\pi_1$ & 74.3 & 15.6 \\
& \method \texttt{KTO} ($\lambda_U:0.0$) $\pi_1$ & 87.1 & 11.6 \\
& \method \texttt{SFT} $\pi_1$ & 96.4 & 10.5 \\
\midrule
\multirow{4}{*}{\rotatebox[origin=c]{90}{out-dist.}} 
& \method $\pi_0$ & 65.7 & 18.6 \\
& \method \texttt{DPO} ($\beta:0.1$) $\pi_1$ & 71.6 & 17.9 \\
& \method \texttt{DPO} ($\beta:0.01$)  $\pi_1$ & 70.9 & 17.7 \\
& \method \texttt{KTO} ($\lambda_U:0.0$) $\pi_1$ & 88.1 & 11.7 \\
& \method \texttt{SFT} $\pi_1$ & 91.0 & 11.9 \\
\bottomrule
\end{tabular}
}
\caption{\textbf{SFT vs Preference:} \method with SFT vs DPO vs KTO on ALFWorld. \figGap
}
\label{tab:sft_vs_preference}
\end{wraptable}


We study the effect of different update methods for \method: SFT vs Preference Optimization. Our initial hypothesis is that KL regularized preference optimization should also result in performance improvement without needing to aggregate datasets. We run $1$ iteration of DPO with different regularization $\beta$ values. Table.~\ref{tab:sft_vs_preference} shows that while  DPO improves upon the base $\pi_0$ policy, the improvement is far smaller than SFT for both $\beta=0.1, 0.01$. One explanation is that SFT is far more aggressive in using the correction data compared to preference optimization, i.e., SFT deems the corrected reason action to be better than any alternative. To test this, we used a different preference optimizer KTO, that operates with unpaired preference data and set weight $\lambda_D=1.0, \lambda_D=0.0$, effectively emulating SFT-like behavior. KTO with these settings has a significantly high performance of $88.1\%$ close to SFT ($91.0\%$).

\vspace{-0.5em}
\section{Related Work}
\vspace{-1em}

\textbf{Imitation Learning and Privileged Information.}
A powerful paradigm in machine learning is leveraging privileged information~\citep{vapnik2015learning}—data available only during training and inaccessible at test time. This concept has been transformative in robotics, where a simulated expert policy with full state supervises a learner that only perceives sensor observations, such as navigation~\citep{zhang2016learning, uppal2024spin}, self-driving~\citep{chen2020learning}, manipulation~\citep{chen2023visual, hu2024privileged}, and legged locomotion~\citep{lee2020learning, kumar2020conservative}.

A fundamental challenge in utilizing privileged information is the realizability gap—where actions suggested by an expert may be infeasible for the learner to predict. This gap often leads to spurious correlations, manifesting as the ``latching effect'' where learners repetitively predict the same action, commonly observed in autonomous driving~\citep{muller2005off, kuefler2017imitating, bansal2018chauffeurnet, codevilla2019exploring} and language models~\citep{bengio2015scheduled, holtzman2019curious, wang2020exposure, ortega2021shaking}. While off-policy methods~\citep{wen2020fighting, wen2022fighting} have been proposed, \citep{swamy2022sequence} show that online interaction with the environment is necessary.

Interactive imitation learning methods such as \DAgger~\citep{ross2011reduction}, where experts provide corrective actions during the learner's rollouts, both theoretically and empirically lead to very effective policies in this regime~\citep{choudhury2018data}. These methods operate under the assumption of asymptotic realizability~\citep{swamy2022sequence}, where the learner's policy can eventually match the expert's actions as the episode progresses. However, when this assumption fails—often due to the partial observability of the task or model capacity—suboptimal performance ensues, particularly for LLM agents, as demonstrated by our experiments (Sec.~\ref{sec:results:tradeoff}). While recent approaches resort to exploration or RL~\citep{walsman2022impossibly, tennenholtz2021covariate, nguyen2022leveraging, weihs2021bridging}, these are intractable to extend to LLM agents. Our work provides a practical recipe for LLM agents that optimally balances the trade-off between using privileged information and maintaining realizability, and further shows how such an expert can be used to enable policy self-improvement.

\textbf{Learning from AI Feedback.} A scalable method for aligning language model using AI feedback~\citep{lee2023rlaif}. Recent works have explored different types of feedback, from self-generated corrections to those provided by external AI models or environments~\citep{pan2023automatically}.
One body of work focuses on self-correction, where LLMs refine their responses based on feedback from their own outputs~\citep{bai2022constitutional, madaan2024self}. This can occur either at train-time, where LLMs are fine-tuned to generate corrected responses~\citep{bai2022constitutional, ganguli2023capacity}, or at test-time, where the model improves its response interactively~\citep{madaan2024self}. However, while simple in principle,~\citep{huang2024llmselfcorrect} shows that without external sources of feedback, LLMs often struggle to accurately judge the correctness of their reasoning, limiting the effectiveness of purely self-generated corrections.

Another cluster of work explores feedback from external sources. Some approaches use feedback from strong models like GPT-4 acting as judges~\citep{zheng2023judging} or critics trained on high-quality curated data~\citep{wang2023shepherd, paul2023refiner}. Others leverage environmental feedback or external tools to guide improvements. \citep{welleck2023generating} uses environmental feedback to train a corrector model, \citep{chen2023teaching, olausson2023demystifying} utilize execution feedback to improve code generation, and \citep{gou2023critic} employs external tools. Notably, Reflexion~\citep{shinn2023reflexion} leverages feedback from external environments to generate self-reflections for agents, although it assumes multiple attempts in the same environment, which our approach does not. Our work introduces the use of privileged information in decision-making tasks for LLM agents, a technique not yet fully explored in the literature. This also enables LLM agents to self-improve using privileged information, setting it apart from existing self-improvement methods that rely on iterative solution generation and evaluation~\citep{zelikman2022star, pang2024iterative, hosseini2024v}.

\vspace{-1em}
\section{Limitations}
\vspace{-1em}

We proposed \method, an iterative fine-tuning framework that improves LLM agent performance using privileged AI feedback, enabling weaker models to surpass stronger expert teachers and allowing agents to self-improve. However, \method has notable limitations. First, generating interaction rollouts is time-consuming, particularly in complex environments requiring multi-step reasoning. Each rollout requires simulating the environment and generating complete trajectories, that becomes computationally expensive across multiple fine-tuning iterations. Second, corrective feedback from the expert often results in lengthy output tokens. As trajectory lengths grow, balancing the effectiveness of feedback with its verbosity becomes challenging. Exploring methods to distill and summarize expert feedback at critical points is an interesting direction for future work.


\newpage
\section*{Reproducibility Statement}

To ensure the reproducibility of our findings, we have made several efforts, as outlined below. Detailed information and resources are provided in the main paper, appendix, and supplementary materials:
\begin{enumerate}[nosep, leftmargin=0.2in]
\item \textbf{Open source code.} We open source the \method code that includes the Python package implementing \method, along with all prompts, configuration files, and training scripts necessary to reproduce our results.
\item \textbf{Experimental details.} We provide detailed descriptions of our experiments in the main paper (Section~\ref{sec:experiments}), with additional details on the setup, hyperparameters, and prompts found in Appendix~\ref{sec:appendix:experiment_details},~\ref{sec:appendix:prompts}. We use open-source datasets and models in our experiments, with citations and links provided in the main paper.
\item \textbf{Analysis.} The key results of our theoretical analysis are presented in the main paper (Section~\ref{sec:approach:analysis}), with full derivations, proofs, and assumptions detailed in Appendix~\ref{sec:appendix:analysis}.
\end{enumerate}
\bibliography{iclr2025_conference}

\begin{thebibliography}{57}
\providecommand{\natexlab}[1]{#1}
\providecommand{\url}[1]{\texttt{#1}}
\expandafter\ifx\csname urlstyle\endcsname\relax
  \providecommand{\doi}[1]{doi: #1}\else
  \providecommand{\doi}{doi: \begingroup \urlstyle{rm}\Url}\fi

\bibitem[Bai et~al.(2022)Bai, Kadavath, Kundu, Askell, Kernion, Jones, Chen,
  Goldie, Mirhoseini, McKinnon, et~al.]{bai2022constitutional}
Yuntao Bai, Saurav Kadavath, Sandipan Kundu, Amanda Askell, Jackson Kernion,
  Andy Jones, Anna Chen, Anna Goldie, Azalia Mirhoseini, Cameron McKinnon,
  et~al.
\newblock Constitutional ai: Harmlessness from ai feedback.
\newblock \emph{arXiv preprint arXiv:2212.08073}, 2022.

\bibitem[Bansal et~al.(2018)Bansal, Krizhevsky, and
  Ogale]{bansal2018chauffeurnet}
Mayank Bansal, Alex Krizhevsky, and Abhijit Ogale.
\newblock Chauffeurnet: Learning to drive by imitating the best and
  synthesizing the worst.
\newblock \emph{arXiv preprint arXiv:1812.03079}, 2018.

\bibitem[Bengio et~al.(2015)Bengio, Vinyals, Jaitly, and
  Shazeer]{bengio2015scheduled}
Samy Bengio, Oriol Vinyals, Navdeep Jaitly, and Noam Shazeer.
\newblock Scheduled sampling for sequence prediction with recurrent neural
  networks.
\newblock \emph{Advances in neural information processing systems}, 28, 2015.

\bibitem[Chen et~al.(2020)Chen, Zhou, Koltun, and
  Kr{\"a}henb{\"u}hl]{chen2020learning}
Dian Chen, Brady Zhou, Vladlen Koltun, and Philipp Kr{\"a}henb{\"u}hl.
\newblock Learning by cheating.
\newblock In \emph{Conference on Robot Learning}, pp.\  66--75. PMLR, 2020.

\bibitem[Chen et~al.(2023{\natexlab{a}})Chen, Tippur, Wu, Kumar, Adelson, and
  Agrawal]{chen2023visual}
Tao Chen, Megha Tippur, Siyang Wu, Vikash Kumar, Edward Adelson, and Pulkit
  Agrawal.
\newblock Visual dexterity: In-hand reorientation of novel and complex object
  shapes.
\newblock \emph{Science Robotics}, 8\penalty0 (84):\penalty0 eadc9244,
  2023{\natexlab{a}}.

\bibitem[Chen et~al.(2023{\natexlab{b}})Chen, Lin, Sch{\"a}rli, and
  Zhou]{chen2023teaching}
Xinyun Chen, Maxwell Lin, Nathanael Sch{\"a}rli, and Denny Zhou.
\newblock Teaching large language models to self-debug.
\newblock \emph{arXiv preprint arXiv:2304.05128}, 2023{\natexlab{b}}.

\bibitem[Choudhury et~al.(2018)Choudhury, Bhardwaj, Arora, Kapoor, Ranade,
  Scherer, and Dey]{choudhury2018data}
Sanjiban Choudhury, Mohak Bhardwaj, Sankalp Arora, Ashish Kapoor, Gireeja
  Ranade, Sebastian Scherer, and Debadeepta Dey.
\newblock Data-driven planning via imitation learning.
\newblock \emph{The International Journal of Robotics Research}, 37\penalty0
  (13-14):\penalty0 1632--1672, 2018.

\bibitem[Codevilla et~al.(2019)Codevilla, Santana, L{\'o}pez, and
  Gaidon]{codevilla2019exploring}
Felipe Codevilla, Eder Santana, Antonio~M L{\'o}pez, and Adrien Gaidon.
\newblock Exploring the limitations of behavior cloning for autonomous driving.
\newblock In \emph{Proceedings of the IEEE/CVF international conference on
  computer vision}, pp.\  9329--9338, 2019.

\bibitem[Dubey et~al.(2024)Dubey, Jauhri, Pandey, Kadian, Al-Dahle, Letman,
  Mathur, Schelten, Yang, Fan, et~al.]{dubey2024llama}
Abhimanyu Dubey, Abhinav Jauhri, Abhinav Pandey, Abhishek Kadian, Ahmad
  Al-Dahle, Aiesha Letman, Akhil Mathur, Alan Schelten, Amy Yang, Angela Fan,
  et~al.
\newblock The llama 3 herd of models.
\newblock \emph{arXiv preprint arXiv:2407.21783}, 2024.

\bibitem[Ethayarajh et~al.(2024)Ethayarajh, Xu, Muennighoff, Jurafsky, and
  Kiela]{ethayarajh2024kto}
Kawin Ethayarajh, Winnie Xu, Niklas Muennighoff, Dan Jurafsky, and Douwe Kiela.
\newblock Kto: Model alignment as prospect theoretic optimization.
\newblock \emph{arXiv preprint arXiv:2402.01306}, 2024.

\bibitem[Ganguli et~al.(2023)Ganguli, Askell, Schiefer, Liao,
  Luko{\v{s}}i{\=u}t{\.e}, Chen, Goldie, Mirhoseini, Olsson, Hernandez,
  et~al.]{ganguli2023capacity}
Deep Ganguli, Amanda Askell, Nicholas Schiefer, Thomas~I Liao, Kamil{\.e}
  Luko{\v{s}}i{\=u}t{\.e}, Anna Chen, Anna Goldie, Azalia Mirhoseini, Catherine
  Olsson, Danny Hernandez, et~al.
\newblock The capacity for moral self-correction in large language models.
\newblock \emph{arXiv preprint arXiv:2302.07459}, 2023.

\bibitem[Gou et~al.(2023)Gou, Shao, Gong, Shen, Yang, Duan, and
  Chen]{gou2023critic}
Zhibin Gou, Zhihong Shao, Yeyun Gong, Yelong Shen, Yujiu Yang, Nan Duan, and
  Weizhu Chen.
\newblock Critic: Large language models can self-correct with tool-interactive
  critiquing.
\newblock \emph{arXiv preprint arXiv:2305.11738}, 2023.

\bibitem[Holtzman et~al.(2019)Holtzman, Buys, Du, Forbes, and
  Choi]{holtzman2019curious}
Ari Holtzman, Jan Buys, Li~Du, Maxwell Forbes, and Yejin Choi.
\newblock The curious case of neural text degeneration.
\newblock \emph{arXiv preprint arXiv:1904.09751}, 2019.

\bibitem[Hosseini et~al.(2024)Hosseini, Yuan, Malkin, Courville, Sordoni, and
  Agarwal]{hosseini2024v}
Arian Hosseini, Xingdi Yuan, Nikolay Malkin, Aaron Courville, Alessandro
  Sordoni, and Rishabh Agarwal.
\newblock V-star: Training verifiers for self-taught reasoners.
\newblock \emph{arXiv preprint arXiv:2402.06457}, 2024.

\bibitem[Hu et~al.(2021)Hu, Shen, Wallis, Allen-Zhu, Li, Wang, Wang, and
  Chen]{hu2021lora}
Edward~J Hu, Yelong Shen, Phillip Wallis, Zeyuan Allen-Zhu, Yuanzhi Li, Shean
  Wang, Lu~Wang, and Weizhu Chen.
\newblock Lora: Low-rank adaptation of large language models.
\newblock \emph{arXiv preprint arXiv:2106.09685}, 2021.

\bibitem[Hu et~al.(2024)Hu, Springer, Rybkin, and Jayaraman]{hu2024privileged}
Edward~S Hu, James Springer, Oleh Rybkin, and Dinesh Jayaraman.
\newblock Privileged sensing scaffolds reinforcement learning.
\newblock \emph{arXiv preprint arXiv:2405.14853}, 2024.

\bibitem[Huang et~al.(2024)Huang, Chen, Mishra, Zheng, Yu, Song, and
  Zhou]{huang2024llmselfcorrect}
Jie Huang, Xinyun Chen, Swaroop Mishra, Huaixiu~Steven Zheng, Adams~Wei Yu,
  Xinying Song, and Denny Zhou.
\newblock Large language models cannot self-correct reasoning yet.
\newblock \emph{arXiv preprint arXiv:2310.01798}, 2024.

\bibitem[Jimenez et~al.(2023)Jimenez, Yang, Wettig, Yao, Pei, Press, and
  Narasimhan]{jimenez2023swe}
Carlos~E Jimenez, John Yang, Alexander Wettig, Shunyu Yao, Kexin Pei, Ofir
  Press, and Karthik Narasimhan.
\newblock Swe-bench: Can language models resolve real-world github issues?
\newblock \emph{arXiv preprint arXiv:2310.06770}, 2023.

\bibitem[Kaelbling et~al.(1998)Kaelbling, Littman, and
  Cassandra]{kaelbling1998planning}
Leslie~Pack Kaelbling, Michael~L Littman, and Anthony~R Cassandra.
\newblock Planning and acting in partially observable stochastic domains.
\newblock \emph{Artificial intelligence}, 101\penalty0 (1-2):\penalty0 99--134,
  1998.

\bibitem[Kuefler et~al.(2017)Kuefler, Morton, Wheeler, and
  Kochenderfer]{kuefler2017imitating}
Alex Kuefler, Jeremy Morton, Tim Wheeler, and Mykel Kochenderfer.
\newblock Imitating driver behavior with generative adversarial networks.
\newblock In \emph{2017 IEEE intelligent vehicles symposium (IV)}, pp.\
  204--211. IEEE, 2017.

\bibitem[Kumar et~al.(2020)Kumar, Zhou, Tucker, and
  Levine]{kumar2020conservative}
Aviral Kumar, Aurick Zhou, George Tucker, and Sergey Levine.
\newblock Conservative q-learning for offline reinforcement learning.
\newblock \emph{Advances in Neural Information Processing Systems},
  33:\penalty0 1179--1191, 2020.

\bibitem[Lee et~al.(2023)Lee, Phatale, Mansoor, Mesnard, Ferret, Lu, Bishop,
  Hall, Carbune, Rastogi, et~al.]{lee2023rlaif}
Harrison Lee, Samrat Phatale, Hassan Mansoor, Thomas Mesnard, Johan Ferret,
  Kellie Lu, Colton Bishop, Ethan Hall, Victor Carbune, Abhinav Rastogi, et~al.
\newblock Rlaif: Scaling reinforcement learning from human feedback with ai
  feedback.
\newblock \emph{arXiv preprint arXiv:2309.00267}, 2023.

\bibitem[Lee et~al.(2020)Lee, Hwangbo, Wellhausen, Koltun, and
  Hutter]{lee2020learning}
Joonho Lee, Jemin Hwangbo, Lorenz Wellhausen, Vladlen Koltun, and Marco Hutter.
\newblock Learning quadrupedal locomotion over challenging terrain.
\newblock \emph{Science robotics}, 5\penalty0 (47):\penalty0 eabc5986, 2020.

\bibitem[Lin et~al.(2018)Lin, Wang, Zettlemoyer, and Ernst]{lin2018nl2bash}
Xi~Victoria Lin, Chenglong Wang, Luke Zettlemoyer, and Michael~D Ernst.
\newblock Nl2bash: A corpus and semantic parser for natural language interface
  to the linux operating system.
\newblock \emph{arXiv preprint arXiv:1802.08979}, 2018.

\bibitem[Madaan et~al.(2024)Madaan, Tandon, Gupta, Hallinan, Gao, Wiegreffe,
  Alon, Dziri, Prabhumoye, Yang, et~al.]{madaan2024self}
Aman Madaan, Niket Tandon, Prakhar Gupta, Skyler Hallinan, Luyu Gao, Sarah
  Wiegreffe, Uri Alon, Nouha Dziri, Shrimai Prabhumoye, Yiming Yang, et~al.
\newblock Self-refine: Iterative refinement with self-feedback.
\newblock \emph{Advances in Neural Information Processing Systems}, 36, 2024.

\bibitem[Muller et~al.(2005)Muller, Ben, Cosatto, Flepp, and
  Cun]{muller2005off}
Urs Muller, Jan Ben, Eric Cosatto, Beat Flepp, and Yann Cun.
\newblock Off-road obstacle avoidance through end-to-end learning.
\newblock \emph{Advances in neural information processing systems}, 18, 2005.

\bibitem[Nguyen et~al.(2022)Nguyen, Baisero, Wang, Amato, and
  Platt]{nguyen2022leveraging}
Hai Nguyen, Andrea Baisero, Dian Wang, Christopher Amato, and Robert Platt.
\newblock Leveraging fully observable policies for learning under partial
  observability.
\newblock \emph{arXiv preprint arXiv:2211.01991}, 2022.

\bibitem[Olausson et~al.(2023)Olausson, Inala, Wang, Gao, and
  Solar-Lezama]{olausson2023demystifying}
Theo~X Olausson, Jeevana~Priya Inala, Chenglong Wang, Jianfeng Gao, and Armando
  Solar-Lezama.
\newblock Demystifying gpt self-repair for code generation.
\newblock \emph{arXiv preprint arXiv:2306.09896}, 2023.

\bibitem[Ortega et~al.(2021)Ortega, Kunesch, Del{\'e}tang, Genewein, Grau-Moya,
  Veness, Buchli, Degrave, Piot, Perolat, et~al.]{ortega2021shaking}
Pedro~A Ortega, Markus Kunesch, Gr{\'e}goire Del{\'e}tang, Tim Genewein, Jordi
  Grau-Moya, Joel Veness, Jonas Buchli, Jonas Degrave, Bilal Piot, Julien
  Perolat, et~al.
\newblock Shaking the foundations: delusions in sequence models for interaction
  and control.
\newblock \emph{arXiv preprint arXiv:2110.10819}, 2021.

\bibitem[Pan et~al.(2023)Pan, Saxon, Xu, Nathani, Wang, and
  Wang]{pan2023automatically}
Liangming Pan, Michael Saxon, Wenda Xu, Deepak Nathani, Xinyi Wang, and
  William~Yang Wang.
\newblock Automatically correcting large language models: Surveying the
  landscape of diverse self-correction strategies.
\newblock \emph{arXiv preprint arXiv:2308.03188}, 2023.

\bibitem[Pang et~al.(2024)Pang, Yuan, Cho, He, Sukhbaatar, and
  Weston]{pang2024iterative}
Richard~Yuanzhe Pang, Weizhe Yuan, Kyunghyun Cho, He~He, Sainbayar Sukhbaatar,
  and Jason Weston.
\newblock Iterative reasoning preference optimization.
\newblock \emph{arXiv preprint arXiv:2404.19733}, 2024.

\bibitem[Papadimitriou \& Tsitsiklis(1987)Papadimitriou and
  Tsitsiklis]{papadimitriou1987complexity}
Christos~H Papadimitriou and John~N Tsitsiklis.
\newblock The complexity of markov decision processes.
\newblock \emph{Mathematics of operations research}, 12\penalty0 (3):\penalty0
  441--450, 1987.

\bibitem[Paul et~al.(2023)Paul, Ismayilzada, Peyrard, Borges, Bosselut, West,
  and Faltings]{paul2023refiner}
Debjit Paul, Mete Ismayilzada, Maxime Peyrard, Beatriz Borges, Antoine
  Bosselut, Robert West, and Boi Faltings.
\newblock Refiner: Reasoning feedback on intermediate representations.
\newblock \emph{arXiv preprint arXiv:2304.01904}, 2023.

\bibitem[Rafailov et~al.(2024)Rafailov, Sharma, Mitchell, Manning, Ermon, and
  Finn]{rafailov2024direct}
Rafael Rafailov, Archit Sharma, Eric Mitchell, Christopher~D Manning, Stefano
  Ermon, and Chelsea Finn.
\newblock Direct preference optimization: Your language model is secretly a
  reward model.
\newblock \emph{Advances in Neural Information Processing Systems}, 36, 2024.

\bibitem[Ross et~al.(2011)Ross, Gordon, and Bagnell]{ross2011reduction}
St{\'e}phane Ross, Geoffrey Gordon, and J~Andrew Bagnell.
\newblock A reduction of imitation learning and structured prediction to
  no-regret online learning.
\newblock In \emph{Artificial Intelligence and Statistics (AISTATS)}, 2011.

\bibitem[Roziere et~al.(2023)Roziere, Gehring, Gloeckle, Sootla, Gat, Tan, Adi,
  Liu, Sauvestre, Remez, et~al.]{roziere2023code}
Baptiste Roziere, Jonas Gehring, Fabian Gloeckle, Sten Sootla, Itai Gat,
  Xiaoqing~Ellen Tan, Yossi Adi, Jingyu Liu, Romain Sauvestre, Tal Remez,
  et~al.
\newblock Code llama: Open foundation models for code.
\newblock \emph{arXiv preprint arXiv:2308.12950}, 2023.

\bibitem[Shinn et~al.(2023)Shinn, Cassano, Labash, Gopinath, Narasimhan, and
  Yao]{shinn2023reflexion}
Noah Shinn, Federico Cassano, Beck Labash, Ashwin Gopinath, Karthik Narasimhan,
  and Shunyu Yao.
\newblock Reflexion: Language agents with verbal reinforcement learning.(2023).
\newblock \emph{arXiv preprint cs.AI/2303.11366}, 2023.

\bibitem[Shridhar et~al.(2020{\natexlab{a}})Shridhar, Thomason, Gordon, Bisk,
  Han, Mottaghi, Zettlemoyer, and Fox]{shridhar2020alfred}
Mohit Shridhar, Jesse Thomason, Daniel Gordon, Yonatan Bisk, Winson Han,
  Roozbeh Mottaghi, Luke Zettlemoyer, and Dieter Fox.
\newblock Alfred: A benchmark for interpreting grounded instructions for
  everyday tasks.
\newblock In \emph{Proc. {IEEE} Int. Conf. Computer Vision and Pattern
  Recognition}, 2020{\natexlab{a}}.

\bibitem[Shridhar et~al.(2020{\natexlab{b}})Shridhar, Yuan, C{\^o}t{\'e}, Bisk,
  Trischler, and Hausknecht]{shridhar2020alfworld}
Mohit Shridhar, Xingdi Yuan, Marc-Alexandre C{\^o}t{\'e}, Yonatan Bisk, Adam
  Trischler, and Matthew Hausknecht.
\newblock Alfworld: Aligning text and embodied environments for interactive
  learning.
\newblock \emph{arXiv preprint arXiv:2010.03768}, 2020{\natexlab{b}}.

\bibitem[Sodhi et~al.(2024)Sodhi, Branavan, Artzi, and McDonald]{sodhi2024step}
Paloma Sodhi, SRK Branavan, Yoav Artzi, and Ryan McDonald.
\newblock Step: Stacked llm policies for web actions.
\newblock In \emph{Conference on Language Modeling (COLM)}, 2024.

\bibitem[Swamy et~al.(2022)Swamy, Choudhury, Bagnell, and
  Wu]{swamy2022sequence}
Gokul Swamy, Sanjiban Choudhury, J~Bagnell, and Steven~Z Wu.
\newblock Sequence model imitation learning with unobserved contexts.
\newblock \emph{Advances in Neural Information Processing Systems},
  35:\penalty0 17665--17676, 2022.

\bibitem[Tennenholtz et~al.(2021)Tennenholtz, Hallak, Dalal, Mannor, Chechik,
  and Shalit]{tennenholtz2021covariate}
Guy Tennenholtz, Assaf Hallak, Gal Dalal, Shie Mannor, Gal Chechik, and Uri
  Shalit.
\newblock On covariate shift of latent confounders in imitation and
  reinforcement learning.
\newblock \emph{arXiv preprint arXiv:2110.06539}, 2021.

\bibitem[Uppal et~al.(2024)Uppal, Agarwal, Xiong, Shaw, and
  Pathak]{uppal2024spin}
Shagun Uppal, Ananye Agarwal, Haoyu Xiong, Kenneth Shaw, and Deepak Pathak.
\newblock Spin: Simultaneous perception interaction and navigation.
\newblock In \emph{Proceedings of the IEEE/CVF Conference on Computer Vision
  and Pattern Recognition}, pp.\  18133--18142, 2024.

\bibitem[Vapnik et~al.(2015)Vapnik, Izmailov, et~al.]{vapnik2015learning}
Vladimir Vapnik, Rauf Izmailov, et~al.
\newblock Learning using privileged information: similarity control and
  knowledge transfer.
\newblock \emph{J. Mach. Learn. Res.}, 16\penalty0 (1):\penalty0 2023--2049,
  2015.

\bibitem[Walsman et~al.(2022)Walsman, Zhang, Choudhury, Fox, and
  Farhadi]{walsman2022impossibly}
Aaron Walsman, Muru Zhang, Sanjiban Choudhury, Dieter Fox, and Ali Farhadi.
\newblock Impossibly good experts and how to follow them.
\newblock In \emph{The Eleventh International Conference on Learning
  Representations}, 2022.

\bibitem[Wang \& Sennrich(2020)Wang and Sennrich]{wang2020exposure}
Chaojun Wang and Rico Sennrich.
\newblock On exposure bias, hallucination and domain shift in neural machine
  translation.
\newblock \emph{arXiv preprint arXiv:2005.03642}, 2020.

\bibitem[Wang et~al.(2023)Wang, Yu, Tan, O'Brien, Pasunuru, Dwivedi-Yu,
  Golovneva, Zettlemoyer, Fazel-Zarandi, and Celikyilmaz]{wang2023shepherd}
Tianlu Wang, Ping Yu, Xiaoqing~Ellen Tan, Sean O'Brien, Ramakanth Pasunuru,
  Jane Dwivedi-Yu, Olga Golovneva, Luke Zettlemoyer, Maryam Fazel-Zarandi, and
  Asli Celikyilmaz.
\newblock Shepherd: A critic for language model generation.
\newblock \emph{arXiv preprint arXiv:2308.04592}, 2023.

\bibitem[Weihs et~al.(2021)Weihs, Jain, Liu, Salvador, Lazebnik, Kembhavi, and
  Schwing]{weihs2021bridging}
Luca Weihs, Unnat Jain, Iou-Jen Liu, Jordi Salvador, Svetlana Lazebnik,
  Aniruddha Kembhavi, and Alex Schwing.
\newblock Bridging the imitation gap by adaptive insubordination.
\newblock \emph{Advances in Neural Information Processing Systems},
  34:\penalty0 19134--19146, 2021.

\bibitem[Welleck et~al.(2023)Welleck, Lu, West, Brahman, Shen, Khashabi, and
  Choi]{welleck2023generating}
Sean Welleck, Ximing Lu, Peter West, Faeze Brahman, Tianxiao Shen, Daniel
  Khashabi, and Yejin Choi.
\newblock Generating sequences by learning to self-correct.
\newblock In \emph{The Eleventh International Conference on Learning
  Representations}, 2023.

\bibitem[Wen et~al.(2020)Wen, Lin, Darrell, Jayaraman, and
  Gao]{wen2020fighting}
Chuan Wen, Jierui Lin, Trevor Darrell, Dinesh Jayaraman, and Yang Gao.
\newblock Fighting copycat agents in behavioral cloning from observation
  histories.
\newblock \emph{Advances in Neural Information Processing Systems},
  33:\penalty0 2564--2575, 2020.

\bibitem[Wen et~al.(2022)Wen, Qian, Lin, Teng, Jayaraman, and
  Gao]{wen2022fighting}
Chuan Wen, Jianing Qian, Jierui Lin, Jiaye Teng, Dinesh Jayaraman, and Yang
  Gao.
\newblock Fighting fire with fire: avoiding dnn shortcuts through priming.
\newblock In \emph{International Conference on Machine Learning}, pp.\
  23723--23750. PMLR, 2022.

\bibitem[Yang et~al.(2024)Yang, Prabhakar, Narasimhan, and
  Yao]{yang2024intercode}
John Yang, Akshara Prabhakar, Karthik Narasimhan, and Shunyu Yao.
\newblock Intercode: Standardizing and benchmarking interactive coding with
  execution feedback.
\newblock \emph{Advances in Neural Information Processing Systems}, 36, 2024.

\bibitem[Yao et~al.(2022{\natexlab{a}})Yao, Chen, Yang, and
  Narasimhan]{yao2022webshop}
Shunyu Yao, Howard Chen, John Yang, and Karthik Narasimhan.
\newblock Webshop: Towards scalable real-world web interaction with grounded
  language agents.
\newblock In \emph{Advances in Neural Information Processing Systems
  (NeurIPS)}, 2022{\natexlab{a}}.

\bibitem[Yao et~al.(2022{\natexlab{b}})Yao, Zhao, Yu, Du, Shafran, Narasimhan,
  and Cao]{yao2022react}
Shunyu Yao, Jeffrey Zhao, Dian Yu, Nan Du, Izhak Shafran, Karthik Narasimhan,
  and Yuan Cao.
\newblock React: Synergizing reasoning and acting in language models.
\newblock \emph{arXiv preprint arXiv:2210.03629}, 2022{\natexlab{b}}.

\bibitem[Zelikman et~al.(2022)Zelikman, Wu, Mu, and Goodman]{zelikman2022star}
Eric Zelikman, Yuhuai Wu, Jesse Mu, and Noah Goodman.
\newblock Star: Bootstrapping reasoning with reasoning.
\newblock \emph{Advances in Neural Information Processing Systems},
  35:\penalty0 15476--15488, 2022.

\bibitem[Zhang et~al.(2016)Zhang, Kahn, Levine, and Abbeel]{zhang2016learning}
Tianhao Zhang, Gregory Kahn, Sergey Levine, and Pieter Abbeel.
\newblock Learning deep control policies for autonomous aerial vehicles with
  mpc-guided policy search.
\newblock In \emph{2016 IEEE international conference on robotics and
  automation (ICRA)}, pp.\  528--535. IEEE, 2016.

\bibitem[Zheng et~al.(2023)Zheng, Chiang, Sheng, Zhuang, Wu, Zhuang, Lin, Li,
  Li, Xing, et~al.]{zheng2023judging}
Lianmin Zheng, Wei-Lin Chiang, Ying Sheng, Siyuan Zhuang, Zhanghao Wu, Yonghao
  Zhuang, Zi~Lin, Zhuohan Li, Dacheng Li, Eric Xing, et~al.
\newblock Judging llm-as-a-judge with mt-bench and chatbot arena.
\newblock \emph{Advances in Neural Information Processing Systems},
  36:\penalty0 46595--46623, 2023.

\end{thebibliography}
\bibliographystyle{iclr2025_conference}


\newpage
\appendix
\addcontentsline{toc}{section}{Appendix} 
\part{Appendix} 
\parttoc 

\onecolumn
\section{Broader Impacts}
\label{sec:appendix:broader_impacts}

\paragraph{Technological Impact.}
By iteratively fine-tuning LLMs through privileged expert feedback, \method can significantly enhance AI performance across diverse applications, such as virtual assistants and dialogue agents handling complex decision-making tasks. This approach allows LLMs to autonomously refine their decision-making, reducing the need for human intervention, streamlining automation, and improving efficiency in both consumer and enterprise environments.

Additionally, \method’s ability to enable weaker models to outperform stronger ones democratizes AI access, making state-of-the-art models more accessible to organizations with limited computational resources. This also opens new possibilities for improving model generalization in settings where perfect information is unavailable. However, the reliance on privileged feedback during training underscores the importance of further research into model transparency and interpretability. As these models grow more autonomous, ensuring their decision-making remains understandable to users will be critical for building trust and fostering widespread adoption.

\paragraph{Societal Impact.}
Equipping LLMs with iterative self-improvement capabilities unlocks a range of societal benefits. In sectors such as education, customer service, and software development, these AI systems can assist professionals in decision-making, automate routine tasks, and enhance overall productivity. By continuously improving their performance, these models can reduce the cognitive load on workers, allowing them to focus on more complex, creative, and value-added activities.

Despite these advantages, there are ethical concerns that must be addressed. As LLMs become capable of self-improvement, ensuring their trustworthiness and accountability becomes critical. Without proper safeguards, the autonomous refinement of AI could lead to unpredictable behaviors, making it harder to ensure alignment with human values. Furthermore, there is a risk that self-improving AIs could be misused for malicious purposes, such as automating harmful tasks or spreading disinformation. To mitigate these risks, strict ethical guidelines and robust oversight is necessary to ensure deployment of such systems serves societal good and prevents misuse.

\section{Analysis}
\label{sec:appendix:analysis}








\subsection{Trade-off between realizability and privileged information}

We now analyze the performance of a policy imitating an expert with access to privileged information. The performance depends both on the performance of the expert policy and the realizability gap between the learner and the expert. We then derive a constrained privileged expert to better trade off these two terms and provide a practical approximation of this constrained expert. For notational simplicity, we assume that $a_t$ represents both reason and action. We follow the derivation in \citep{swamy2022sequence} but simplify for our setting. 

Let $\pi \in \Pi$ be the student policy that we are training. We assume the class of policies $\Pi$ is rich enough to contain all history-based policies. 
Let $J(\pi)$ be the performance, i.e. the cumulative reward of a learner policy that selects actions based on the history $a_t = \pi(h_t)$. Let $J(\pi_E)$ be the performance of the expert policy that selects actions based on the privileged state $a_t = \pi^E(s_t)$. Let $\tau \sim \pi$ be a \emph{privileged rollout} which is the sequence of observation, actions, and privileged state upon executing the policy. We represent the rollout as $\tau=\{(h_t, s_t, a_t)\}$. Note that such a privileged rollout is only accessible at train time.

We begin by defining the average imitation gap between an expert policy $\pi^E$ and learner $\pi$. 

\begin{definition}[Average Imitaiton Gap] The average imitation gap of a learner $\pi$ over a horizon $T$:
\begin{equation}
    \mathrm{AIG}(\pi, T) := \frac{1}{T} \left( J(\pi^E) - J(\pi) \right) = \mathbb{E}_{\tau \sim \pi^E} \left[\frac{1}{T} \rew(s_t, a_t) \right] - \mathbb{E}_{\tau \sim \pi} \left[ \frac{1}{T}  \rew(s_t, a_t) \right]
\end{equation}
\end{definition}

We next define the realizability gap between the privileged expert and the best policy in policy class as the L1 distance (or equivalently Total-Variation (TV) distance) between the two policies on the distribution of states and histories induced by \emph{any policy}.

\begin{definition}[Average Realizability Gap] The average realizability gap over time between the privileged expert $\pi^E$ and the best policy in the policy class $\pi^\star$ is:
\begin{equation}
\epsilon (\pi^E, T)  := \sup_{\pi} \frac{1}{T} \sum_{t=1}^{T} \mathbb{E}_{s_t, h_t \sim d^\pi_t} || \pi^E(a_t|s_t) - \pi^{\star}(a_t | h_t)) ||_1
\end{equation}
\end{definition}


We next define the recoverability of the expert. Let $A^{\pi^E}(s,a) = Q^{\pi^E}(s,a) - V^{\pi^E}(s)$ be the advantage of a state action $(s,a)$ with respect to the expert policy $\pi^E$. A high negative advantage means that the mistaken action at a state results in a high value difference from the action the expert would have taken. Recoverability is the worst case advantage. 

\begin{definition}[Recoverability Coefficient of the Privileged Expert] The recoverability of the expert policy $\pi^E$ is the max advantage of the expert policy
\begin{equation}
    H(\pi^E) :=  \max_{s,a} || A^{\pi^E} (s,a) ||
\end{equation}
\end{definition}

We are now ready to state the main theorem of \method.

\begin{theorem}[\method with Privileged Expert] Running $N$ iterations of \method with privileged expert $\pi^E$ yields at least one policy $\pi$ such that
\begin{equation}
    \frac{1}{T} J(\pi) \geq \frac{1}{T} J(\pi^E)  - H(\pi^E) \left( \epsilon (\pi^E, T) + \gamma(N) \right)
\end{equation}
where $\pi^E$ is the privileged expert, $H(\pi^E)$ is the recoverability coefficient of $\pi^E$, $\epsilon (\pi^E, T)$ is the average realizability gap, and $\gamma(N)$ is the average regret of the \DAgger update. 
\end{theorem}

The first term is the performance of the privileged expert and the second term is the average realizability of the privileged expert. The more privileged information the expert policy $\pi^E$ leverages the better $J(\pi^E)$ and the lower recoverability coefficient $H(\pi^E)$. However, the realizability gap $\epsilon (\pi^E, T)$ grows larger. 

\begin{proof}
\DAgger~\citep{ross2011reduction} with $\pi^E$ as the expert teacher guarantees that after $N$ iterations, it will find at least one policy
\begin{equation}
\label{eq:dagger:priv}
\mathbb{E}_{s, h \sim d^\pi} ||\pi(.|h) - \pi^E (.|s,h)||_1 \leq \mathbb{E}_{s, h \sim d^\pi} ||\pi^\star(.|h) - \pi^E(.|s,h)||_1 + \gamma(N) 
\end{equation}
where $\gamma(N)$ is the average regret, and $d^\pi$ is the time average distribution of states and history induced by policy $\pi$.

\begin{equation*}
\begin{aligned}
    \mathrm{AIG}(\pi, T) &= \frac{1}{T} \left( J(\pi^E) - J(\pi) \right) \\
    &\leq \frac{1}{T} \sum_{t=1}^T \mathbb{E}_{ s_t, h_t \sim d^\pi_t } \left[ \sum_a A^{\pi^E}(s_t,a_t) \left( \pi^E(a|s_t) - \pi(a|h_t) \right) \right] & \text{(PDL)} \\
     & \leq  ||A^{\pi^E} (., .)||_\infty  \frac{1}{T} \sum_{t=1}^T \mathbb{E}_{ s_t, h_t \sim d^\pi_t } ||\pi(.|h_t) - \pi^E(.|s_t, h_t)||_1 & \text{(Holder's Inequality)} \\
     & \leq  H(\pi^E)  \frac{1}{T} \sum_{t=1}^T \mathbb{E}_{ s_t, h_t \sim d^\pi_t } ||\pi(.|h_t) - \pi^E(.|s_t, h_t)||_1 \\
     & \leq  H(\pi^E)  \left( \frac{1}{T} \sum_{t=1}^T \mathbb{E}_{ s_t, h_t \sim d^\pi_t } ||\pi^\star(.|h_t) - \pi^E(.|s_t, h_t)||_1 + \gamma(N) \right) & \text{From (\ref{eq:dagger:priv}) }\\
     & \leq  H(\pi^E) \left( \epsilon (\pi^E, T) + \gamma(N) \right)\\
\end{aligned}
\end{equation*}

Rearranging, we get
\begin{equation*}
    \frac{1}{T} J(\pi) \geq \frac{1}{T} J(\pi^E)  - H(\pi^E) \left( \epsilon (\pi^E, T) + \gamma(N) \right)
\end{equation*}
\end{proof}

To optimize the performance bound, we introduce a constrained privileged expert that provides actions similar to the privileged expert but is constrained to be close to realizable. 

\begin{definition}[Constrained Privileged Expert] 
\label{eq:constrained_priv_expert}
Given a state $s_t$ and history $h_t$, a constrained privileged expert minimizes KL distance to the privileged expert while being in a $\delta$ KL ball of the non-privileged expert.
\begin{equation}
\label{eq:constrained:expert}
    \pi^E_{\delta}(.|s_t, h_t) := \arg\min_{\tilde{\pi}} \mathrm{KL}\left( \tilde{\pi}(.|s_t, h_t) || \pi^E(.|s_t) \right) \quad \mathrm{s.t.} \quad \mathrm{KL}\left( \tilde{\pi}(.|s_t, h_t) || \pi^E(.|h_t) \right) \leq \delta 
\end{equation}
\end{definition}

We use the fact that we have a rich policy class $\Pi$ that includes the non-privileged expert $\pi^E(.|h)$, to show that the constrained privileged expert bounds the average realizability gap
\begin{equation}
\label{eq:cons_realizability_gap}
\begin{aligned}
    \epsilon (\pi^E_\delta, T)  &= \sup_{\pi} \frac{1}{T} \sum_{t=1}^{T} \mathbb{E}_{s_t, h_t \sim d^\pi_t} || \pi^E_\delta(a_t|s_t) - \pi^{\star}(a_t | h_t)) ||_1 \\
    &\leq \sup_{\pi} \frac{1}{T} \sum_{t=1}^{T} \mathbb{E}_{s_t, h_t \sim d^\pi_t} \sqrt{2 \mathrm{KL}( \pi^E_\delta(a_t|s_t) || \pi^{\star}(a_t | h_t) )} \\
    &\leq \sup_{\pi} \frac{1}{T} \sum_{t=1}^{T} \mathbb{E}_{s_t, h_t \sim d^\pi_t} \sqrt{2 \mathrm{KL}( \pi^E_\delta(a_t|s_t) || \pi^{E}(a_t | h_t) )} & \text{ since $\pi^{E}(.| h_t) \in \Pi$ }\\
    &\leq \sqrt{2 \delta} & \text{ from (\ref{eq:constrained:expert}) }
\end{aligned}
\end{equation}

\begin{theorem} [\method with Constrained Privileged Expert] Running $N$ iterations of \method with constrained privileged expert $\pi^E_\delta$ yields at least one policy $\pi$ such that
\begin{equation}
    \frac{1}{T} J(\pi) \geq \frac{1}{T} J(\pi^E)  - \left( \frac{1}{T} \underbrace{\left( J(\pi^E) - J(\pi^E_\delta) \right)}_{\textup{gap from privileged expert}} + H(\pi^E_\delta) \left( \underbrace{ \sqrt{2\delta}}_{\textup{realizability}} + \gamma(N)\right) \right)
\end{equation}
where $\pi^E$ is the privileged expert, $H(\pi^E_\delta)$ is the recoverability coefficient of $\pi^E_\delta$, $\delta$ is the KL constraint, $\gamma(N)$ is the average regret of the \DAgger update. 
\end{theorem}

The first term is the performance gap between the privileged and the constrained privileged expert. The second term is the realizability of the constrained privileged expert. As $\delta \rightarrow 0$, the realizability gap goes down while the performance gap and recoverability coefficient go up. There is a critical value of $\delta$ that achieves an optimal tradeoff. 
\begin{proof}

\DAgger~\citep{ross2011reduction} with $\pi^E_\delta$ as the expert teacher guarantees that after $N$ iterations, it will find at least one policy
\begin{equation}
\label{eq:dagger:cons}
\mathbb{E}_{s, h \sim d^\pi} ||\pi(.|h) - \pi^E_\delta (.|s,h)||_1 \leq \mathbb{E}_{s, h \sim d^\pi} ||\pi^\star(.|h) - \pi^E_\delta(.|s,h)||_1 + \gamma(N) 
\end{equation}
where $\gamma(N)$ is the average regret. 

\begin{equation*}
\begin{aligned}
    &\mathrm{AIG}(\pi, T) = \frac{1}{T} \left( J(\pi^E) - J(\pi) \right) \\
    &= \frac{1}{T} \left(J(\pi^E) - J(\pi^E_\delta)\right) + \frac{1}{T} \left( J(\pi_\delta^E) - J(\pi) \right) \\
    &\leq \frac{1}{T} \left(J(\pi^E) - J(\pi^E_\delta)\right) + \frac{1}{T} \sum_{t=1}^T \mathbb{E}_{ s_t, h_t \sim d^\pi_t } \left[ \sum_a A^{\pi^E_\delta}(s_t,a_t) \left( \pi^E_\delta(a|s_t) - \pi(a|h_t) \right) \right] & \text{(PDL)} \\
     & \leq \frac{1}{T} \left(J(\pi^E) - J(\pi^E_\delta)\right) + ||A^{\pi^E_\delta} (., .)||_\infty  \frac{1}{T} \sum_{t=1}^T \mathbb{E}_{ s_t, h_t \sim d^\pi_t } ||\pi(.|h_t) - \pi^E_\delta(.|s_t, h_t)||_1 & \text{(Holder's Inequality)} \\
     & \leq \frac{1}{T} \left(J(\pi^E) - J(\pi^E_\delta)\right) + H(\pi^E_\delta)  \frac{1}{T} \sum_{t=1}^T \mathbb{E}_{ s_t, h_t \sim d^\pi_t } ||\pi(.|h_t) - \pi^E_\delta(.|s_t, h_t)||_1 \\
     & \leq \frac{1}{T} \left(J(\pi^E) - J(\pi^E_\delta)\right) + H(\pi^E_\delta)  \left( \frac{1}{T} \sum_{t=1}^T \mathbb{E}_{ s_t, h_t \sim d^\pi_t } ||\pi^\star(.|h_t) - \pi^E_\delta(.|s_t, h_t)||_1 + \gamma(N) \right) & \text{From (\ref{eq:dagger:cons}) }\\
     & \leq \frac{1}{T} \left(J(\pi^E) - J(\pi^E_\delta)\right) + H(\pi^E_\delta) \left( \epsilon (\pi^E_\delta, T) + \gamma(N) \right)\\
    & \leq \frac{1}{T} \left(J(\pi^E) - J(\pi^E_\delta)\right) + H(\pi^E_\delta) \left( \sqrt{2\delta} + \gamma(N) \right) & \text{From (\ref{eq:cons_realizability_gap})}\\
\end{aligned}
\end{equation*}

Rearranging, we get

\begin{equation*}
    \frac{1}{T} J(\pi) \geq \frac{1}{T} J(\pi^E)  - \left( \frac{1}{T} \left( J(\pi^E) - J(\pi^E_\delta) \right) + H(\pi^E_\delta) \left( \sqrt{2\delta} + \gamma(N) \right) \right)
\end{equation*}

\end{proof}

\subsection{Practical approximation for constrained privileged expert}

Writing out the Lagrangian of the constrained privileged expert (Def.~\ref{eq:constrained_priv_expert}) and solving for the primal we get:
\begin{equation}
    \pi^E(\rho, a | s, h) = \frac{1}{Z(s, h)} \pi^E(\rho, a_t|s_t) \left( \pi^E(\rho, a_t|h_t) \right)^\lambda
\end{equation}
where $\lambda$ is the Lagrange multiplier.

For a fixed multiplier $\lambda$, this closed-form expression suggests a simple penalty-method approximation to constrained privileged expert:
\begin{enumerate}
    \item Sample $N$ responses from both  $\pi^E ( \tilde{\rho}, \tilde{a} | {\rm \rho}, {a}, h, s)$ and $\pi^E ( \tilde{\rho}, \tilde{a} | {\rm \rho}, {a}, h)$  to create $\mathcal{R} = \{ \tilde{\rho}^i, \tilde{a}^i \}_{i=1}^N$
    \item Reweigh the responses in $\mathcal{R}$  according to
    \begin{equation}
     \log \pi^E ( \tilde{\rho}, \tilde{a} | {\rho}, {a}, h, s) + \lambda \log \pi^E ( \tilde{\rho}, \tilde{a} | {\rho}, {a}, h) 
    \end{equation}
     \item Sample from the reweighed distribution.
\end{enumerate}

Since the majority of our experiments involve closed-weights models (like \gptfo), we are unable to access the logits of the expert. Instead, we design \method prompts to approximate the constrained privileged expert by providing instructions to not reveal the privileged information in the reasoning, and provide general reason and actions that should be predictable given only history $h_t$.

\section{Experimental details}
\label{sec:appendix:experiment_details}

\subsection{Hyper-parameters}

\subsubsection{Training parameters}
Tables~\ref{tab:training_params_sft} and \ref{tab:training_params_pref} contain hyperparameters for SFT training and DPO/KTO training using LoRA for the different datasets. All training runs were on machines with either 2 or 4 RTX A6000 GPUs, each with 48 GB of memory per GPU.

\begin{table}[H]
\centering
\resizebox{0.8\columnwidth}{!}{
\begin{tabular}{llll}
\toprule
\textbf{Dataset} & \textbf{AlfWorld} & \textbf{WebShop} & \textbf{InterCode} \\
\midrule
Model & Llama3-8B & Llama3-8B & Llama3-8B / Llama3-70B \\
Batch size & 64 & 16 & 16 \\
Max seq length & 8000 & 6000 & 6000 \\
Max epochs & 1 & 1 & 5 \\
Learning rate & 3e-5 & 3e-5 & 3e-5 \\
LoRA $\alpha$ & 64 & 64 & 64 \\
LoRA $r$ & 128 & 128 & 128 \\
LoRA dropout & 0.05 & 0.05 & 0.05 \\
Optimizer & AdamW & AdamW & AdamW \\
LR scheduler & cosine & cosine & cosine \\
\bottomrule
\end{tabular}
}
\caption{Hyperparameters for SFT training using LoRA}
\label{tab:training_params_sft}
\end{table}

\begin{table}[H]
\centering
\resizebox{0.5\columnwidth}{!}{
\begin{tabular}{lll}
\toprule
\textbf{Method} & \textbf{DPO} & \textbf{KTO} \\
\midrule
Model & Llama3-8B & Llama3-8B \\
Batch size & 32 & 64 \\
Max seq length & 8000 & 8000 \\
Max prompt length & 6000 & 6000 \\
Max epochs & 1 & 1 \\
Regularization $\beta$ & 0.01 & 0.01 \\
Learning rate & 5e-7 & 5e-7 \\
LoRA $\alpha$ & 64 & 64 \\
LoRA $r$ & 128 & 128 \\
LoRA dropout & 0.05 & 0.05 \\
Optimizer & AdamW & AdamW \\
LR scheduler & cosine & cosine \\
\bottomrule
\end{tabular}
}
\caption{Hyperparameters for preference training (DPO, KTO) using LoRA}
\label{tab:training_params_pref}
\end{table}

\subsubsection{Evaluation parameters}
For inference using \texttt{Llama3-8b} and \texttt{Llama3-70b}, we use a temperature setting of 0.3 and a maximum token length of 256. The following code snippet shows the \texttt{.generate()} function used:
\begin{lstlisting}[language=Python, caption=Generation function with hyperparameters for Llama3 models]
outputs = model.generate(
    tokenized_inputs["input_ids"],
    attention_mask=tokenized_inputs["attention_mask"],
    max_new_tokens=256,
    eos_token_id=[
        tokenizer.eos_token_id,
        tokenizer.convert_tokens_to_ids("<|eot_id|>"),
    ],
    temperature=0.3,
    pad_token_id=self.tokenizer.eos_token_id
)
\end{lstlisting}

For all OpenAI model calls, including tasks such as generating expert teacher corrections, labeling reasoning for actions, or running prompt baselines like ReAct, we use the API call with the following hyperparameters. See  Appendix~\ref{sec:appendix:prompts} for the detailed prompts.
\begin{lstlisting}[language=Python, caption=Hyperparameters for OpenAI GPT-4o and GPT-4o-mini models]
#  model: gpt-4o
chat_completion = client.chat.completions.create(
    messages=messages,
    model="gpt-4o",
    temperature=0.3, 
    top_p=1, 
    n=1
)
response = chat_completion.choices[0].message.content

#  model: gpt-4o-mini
chat_completion = client.chat.completions.create(
    messages=messages,
    model="gpt-4o-mini",
    temperature=0.3, 
    top_p=1, 
    n=1
)
response = chat_completion.choices[0].message.content
\end{lstlisting}

\subsection{ALFWorld}
\label{sec:appendix:setup:alfworld}

ALFWorld~\citep{shridhar2020alfworld} is a text-based game built on the ALFRED embodied benchmark \citep{shridhar2020alfred}. Each task has a high-level objective, e.g. ``heat mug and put it in cabinet'' which the agent must complete by navigating and interacting with a virtual household through text commands, e.g. go to shelf 1, pick up mug 2, etc. It is a POMDP where the hidden state is the location of all objects, and the observations reveal only the objects the agent can see. To solve it, the agent must plan subgoals, track progress, and explore efficiently till time runs out. An interesting aspect of ALFWorld is the requirement to infer the likely locations of household objects (e.g. mugs are more likely to be on shelves or cabinets), which the agent can learn from interaction with the environment. There are $6$ categories of tasks and two datasets: $139$ in-distribution games and 134 out-of-distribution games. Time horizon is limited to $30$. 

We compare against \texttt{ReAct}~\citep{yao2022react} with different base models and prompts. The original \react paper~\citep{yao2022react} used text-davinci-002 with task-dependent in-context examples. We run \react wtug \gptfo and \gptfomini and an instruction prompt that is shared by all our methods. We also compare against BUTLER~\citep{shridhar2020alfworld}, a seq2seq BERT model trained using DAGGER.\footnote{Since we look at the single trial setting, it rules out approaches like Reflexion~\citep{shinn2023reflexion} where agents require multiple trials on the same game.}

We LoRA fine-tune Llama-3-8B model using \gptfo as the expert teacher for $4$ iterations. We use observation-action demonstrations on the training dataset of $3257$ games\cite{shridhar2020alfworld}, which we annotate with \gptfo to generate reasons and use this to train the BC policy $\pi_0$. We then roll out a policy and generate corrections from the privileged expert on unsuccessful rollouts. The correction data is then aggregated with the existing dataset and used to train a policy using SFT.

The privileged state in ALFworld consists of 
\begin{enumerate}
    \item Task: What is the overall task
    \item Essential Objects: Enumerate all objects that are necessary to solve the game.
    \item Critical Locations: List the important locations involved in the game solution.
    \item Optimal Action Sequence: Suggest an optimal sequence of actions that could theoretically minimize the steps needed to complete the game, based on the log provided. Be sure to copy the action verbatim.
\end{enumerate}

\subsection{WebShop}
\label{sec:appendix:setup:webshop}
WebShop~\citep{yao2022webshop} is an online shopping environment consisting of $1.18$ million real-world products and $12,586$ human-provided instructions. Each task is a shopping request from a user (``I need a gluten-free bacon, 4 oz pack of 2, priced under \$50'') that the agent has to solve through a set of interactions (search[``gluten-free bacon''] or selecting options such as "4 oz pack of 2"). The metrics are the overall score, and 4 components of the score: whether the attributes match (\textsc{Att}), whether the options match (\textsc{Opt}), whether the product types match (\textsc{Type}) and whether the price matches (\textsc{Price}). We evaluate on $500$ test instructions, and limit time horizon to $30$.  

We compare against \texttt{ReAct}~\citep{yao2022react} with different base models like GPT-4o and GPT-4o-mini. We use the same instruction prompt as in \method. We also compare against IL baseline in WebShop~\citep{yao2022webshop} trained using BC. These baselines train two different selection models, one for selecting search queries from a list of options and one for clicking from a list of options. This constraint search space simplifies the problem. 

We LoRA fine-tune Llama-3-8B model using GPT4-o as the expert teacher for 4 iterations. We initialize $\pi_0$ using observation-action demonstrations from \cite{yao2022webshop} which we annotate with GPT4-o to generate reasons. We roll-out policies on the training dataset of $12086$ tasks, and generate corrections from the privileged expert on unsuccessful rollouts. For every game in the training dataset, we construct privileged information that contains: the attributes the product must satisfy, the price constraint, the option that has to be selected, and an example product that would satisfy all criteria. The correction data is then aggregated with the existing dataset and used to train a policy using SFT. 

The privileged state in ALFworld consists of 
\begin{enumerate}
    \item Task: What is the user instruction
    \item Attributes: What are the attributes the product must have
    \item Price: What price constraints must the product satisfy.
    \item Example Product: Name of a product that satisfies all criteria.
\end{enumerate}

\subsection{Intercode}
\label{sec:appendix:setup:intercode}

Intercode~\cite{yang2024intercode} is an interactive coding environment, where an agent interacts with a compiler/interpreter to execute its code and submit further refinements. We use the Bash environment, where a typical task is ``Find all text files and write their names to a single file.'', and the agent issues commands like \texttt{ls}, \texttt{find ..} to solve the task. Intercode builds on the NL2Bash~\citep{lin2018nl2bash} to create $4$ interactive datasets. We use the first $2$ for training, and the next $2$ for testing. We limit the time horizon to $10$. We compare against \react~\cite{yao2022react} with \gptfo and \gptfomini. 
We also compare against the top 3 entries on the Intercode leaderboard, which are the \texttt{TryAgain} baselines from \cite{yang2024intercode} with \gptf, \gptt and CodeLlama-34B-INST respectively. Note the numbers are directly from the leaderboard which evaluates on all (train+test) data and, while our numbers are only on test.

\section{Prompts}
\label{sec:appendix:prompts}

We define the following three prompts across all datasets:
\begin{enumerate}[nosep, leftmargin=0.4in]
    \item \textbf{Expert corrections template:} This prompt is used by the LLM expert teacher to provide corrections on the student agent's trajectories. These corrections offer improved reasoning and actions that balance the trade-off between utilizing the privileged state (accessible only to the expert during training) and ensuring the corrections are realizable by the student agent.
    
    \item \textbf{Student agent template:} This prompt is used during test time by the Llama3 student agent models to generate reason and action trajectories that achieve the task objective. It is also used for prompting baselines, such as GPT-4 or ReAct, which generate reasoning and action trajectories to achieve the task objectives.

    \item \textbf{Autolabel reasons template:} For datasets that contain human demonstrations with action labels (like AlfWorld, WebShop), this prompt is used to generate reasoning annotations for actions that the human took. Generating reasons is necessary as the student agent needs to be trained to produce both reason and actions in its output trajectories.
\end{enumerate}

For prompts used in different ablation studies, please refer to the accompanying code.

\subsection{AlfWorld}
\label{sec:appendix:prompts:alfworld}

\subsubsection{{Expert corrections template}}

\begin{lstlisting}[language=Python]
{% if system %}
You are given a trajectory containing observations, reason and actions generated by a student agent solving a text world game. You are a teacher who has access to a "privileged state" that contains secret information sufficient to solve the game but is hidden from the student. Your goal is to improve how the student solves the game by improving their reason and action at every timestep. 

### Input
You will be provided with a JSON file that logs the student agent's observation, candidate_actions, reason and action at every timestep while playing a text-based game. The student fails to solve the task within the time horizon of the game. 

The structure is an array of objects, containing the following at each timestep:
- timestep: Index of the current timestep
- observation: The observation provided to the student agent
- candidate_actions: The set of allowed actions
- reason: The reason generated by the student agent to justify their action
- action: The action taken by the student agent

You will also be provided the privileged state that contains hidden information that specifies how to solve the task. 

### Task
* Analyze the student trajectory and summarize the mistakes it is making when trying to solve the game
* Refer to your privileged state to know how the game can be solved
* Generate IMPROVED reason and action for the student at every timestep to guide them towards the goal
* Base your improved reasons solely on the student's historical observations and actions up to each timestep
* Do NOT include any information from your privileged state in the improved reasons, as the student does not have access to those
* Offer GENERAL principles or hints in your improved reasons that explain why the student should prefer your suggested action over their original action. This would help the student generalize better.
* When generating improved reason, action at timestep t, assume that the student has followed their original trajectory up until timestep t. Copy over the original observation at timestep t from the student trajectory. 
* When generating improved action at timestep t, make sure the action is available in the candidate_actions at timestep t. Don't select an action that is not available.

Important: Provide GENERAL principles or hints in your improved reasons that explain why the student should prefer your suggested action over their original action. 
* If you know from your privileged state that an object is in a different location from where the student is exploring, use common sense rationale to guide the student
* Do not directly instruct the student to go to the desired location, as they do not have access to the privileged information you possess
* For example, if the student is exploring the wrong area, instead of stating the object is in a different location, suggest a general principle like, "It might be useful to explore areas because ... ", ".. this item is often found in such places", etc
* By following these steps, you help the student understand the logic behind the actions without revealing privileged information
* Also note that you cannot carry more than one item at a time

### Output
The output is a JSON containing a summary and a trajectory with the same length as the input student trajectory as follows:
```json
{
    "summary": your summary of the mistakes the student is making,
    "trajectory": [
    {
        "timestep": Index of the current timestep,
        "original_observation": The original observation made by student agent at timestep t (copy as is from student trajectory),
        "original_reason": The original reason generated by the student at timestep t (copy as is from student trajectory),
        "original_action": The original action the student took at timestep t (copy as is from student trajectory),
        "corrected_reason": The corrected reason that the student should generate,
        "corrected_action": The corrected action that the student should take (chosen from list of candidate_actions at timestep t)
    },
    {
        ...
    }
    ...
    ]
}
```
{% endif %}
{% if not system %}
The student trajectory is below:
{{student_trajectory}}

The privileged state for the task is below:
{{privileged_state}}

Provide the ### Output in the JSON format specified above.
{% endif %}
"""
\end{lstlisting}

\subsubsection{Student agent template}
\begin{lstlisting}[language=Python]
{% if mode == 'input' %}
You are an intelligent assistant named ALFRED in a text-based interactive game called TextWorld. Your objective is to complete the given tasks by reasoning through the information provided and taking appropriate actions. 

Your task is the following:
{{task}}

Below is the history of previous observations and actions:
{{ observation_action_history }}


Given the history of previous observation and action above, a reminder that your task is:
{{task}}

You are given as input the current observation and the list of possible candidate_actions:
{
    "observation": {{observation}},
    "candidate_actions": {{candidate_actions}}
}


Your goal is to generate the action to take at this time step (chosen from candidate_actions) along with the reason for taking the action. 

Please follow these general instructions:
* You MUST choose action from the list of candidate_actions.
* If "observation": "Nothing happens.", it is because you chose an invalid action not from the list of candidate_actions in the previous timestep.
* Oftentimes the task requires you interact with objects not present in your observation. You must search the environment to locate the objective.
* Consult the history of previous observations and actions to see what actions you have tried already so as to not repeat your actions.
* Do NOT repeat the same action as the last action in your observation_action_history. It's going to yield the same result. 
* Make sure action is VERBATIM copied from the list of candidate_actions.

You need to generate a response in the following format. Please issue only a single action at a time.
REASON:
Rationale for what action to take next based on the task and previous history. In your reason, consult candidate_actions to precisely state VERBATIM which action you will do.
ACTION:
The action to be taken, chosen ONLY from candidate_actions
{% elif mode == 'output' %}
REASON:
{{ reason }}
ACTION:
{{ action }}
{% endif %}
\end{lstlisting}

\subsubsection{Autolabel reasons template} 

\begin{lstlisting}[language=Python]
{% if system %}
You are given a JSON file containing observations and actions generated by an expert agent solving text world games. Your goal is to generate the reasoning process of the agent to justify each of their actions.

### Input
The input is a JSON array of objects, each containing:
- `observation`: The state observed by the agent at a specific timestep.
- `action`: The action taken by the agent in response to the observation.

Example input format:
```json
[
    {
        "observation": "You see a locked door.",
        "action": "Use key on door."
    },
    ...
]
```

### Output
The output should be a JSON array of objects, each containing:
- `observation`: The state observed by the agent at a specific timestep.
- `reason`: The reasoning behind the action taken by the agent.
- `action`: The action taken by the agent in response to the observation.

Example output format:
```json
[
    {
        "observation": "You see a locked door.",
        "reason": "The door is locked and I have a key that might open it.",
        "action": "Use key on door."
    },
    ...
]
```

### Requirements
1. **Causal Constraints**: Ensure the reasoning satisfies causal constraints. Use only the observations up to the current timestep to justify the action. Do not use future information.
2. **Brevity and Relevance**: The reasoning should be brief and relevant, using the observations up until that timestep to justify the action.
3. **Sequence Consistency**: Ensure the output maintains the same sequence of observations and actions as in the input.

{% endif %}

{% if not system %}
The input JSON file is below:
{{input}}

Provide the output in the format specified above.
{% endif %}
\end{lstlisting}

\subsection{WebShop}
\label{sec:appendix:prompts:webshop}

\subsubsection{{Expert corrections template}}
\begin{lstlisting}[language=Python]
{% if system %}
You are given a trajectory containing observations, reason and actions generated by a student agent solving a shopping task by searching for items in a website and clicking on links. You are a teacher who has access to a "privileged state" that contains secret information sufficient to solve the task but is hidden from the student. Your goal is to improve how the student solves the task by improving their reason and action at every timestep. 

### Input
You will be provided with a JSON file that logs the student agent's observation, candidate_actions, reason and action at every timestep while solving a shopping task. The student fails to solve the task within the time horizon of the task. 

The structure is an array of objects, containing the following at each timestep:
- timestep: Index of the current timestep
- observation: The observation provided to the student agent
- candidate_actions: The set of allowed actions
- reason: The reason generated by the student agent to justify their action
- action: The action taken by the student agent

You will also be provided the privileged state that contains hidden information that specifies how to solve the task. 

### Task
* Analyze the student trajectory and summarize the main mistakes it is making when trying to solve the task
* Refer to your privileged state to know how the task can be solved
* Correct a select set of timesteps where you know the student was definitely doing a wrong action 
* Generate IMPROVED reason and action for the student at every timestep to guide them towards the goal
* Base your improved reasons solely on the student's historical observations and actions up to each timestep
* Do NOT include any information from your privileged state in the improved reasons, as the student does not have access to those
* Offer GENERAL principles or hints in your improved reasons that explain why the student should prefer your suggested action over their original action. This would help the student generalize better.
* When generating improved reason, action at timestep t, assume that the student has followed their original trajectory up until timestep t. Copy over the original observation at timestep t from the student trajectory. 
* When generating improved action at timestep t, make sure the action is available in the candidate_actions at timestep t. Don't select an action that is not available.  If candidate_actions has only click actions, you can only choose a click action from the list. If candidate_actions has only search[<search query>], you must search by generating the search keywords on your own.

Important: 
(1) Provide GENERAL principles or hints in your improved reasons that explain why the student should prefer your suggested action over their original action. 
* If you know from your privileged state that the desired product is different from the one the student is considering, use common sense rationale to guide the student
* Do not directly instruct the student to search the ground truth product, as they do not have access to the privileged information you possess
* By following these steps, you help the student understand the logic behind the actions without revealing privileged information

(2) Put corrections sparingly. 
* Be strategic about which timesteps you want to correct.
* If you think an action is good enough or reasonable, don't bother correcting. Just copy over the original reason and action.

(3) Correct indecisive behavior of the student agent
* Often times, you will see the student agent not finishing the task and instead continually browsing through items.
* Provide corrections at key moments to help it resolve indecisiveness. 
* It's better to purchase a suboptimal item than not finish the task within the time horizon.
* The student gets partial points for matching a subset of the criteria in the instruction. They get 0 points for not finishing the task in the time horizon (max timesteps in student trajectory)

(4) Considerations for search.
* Specifying prices in search does not work. Other specifications are fine. 
* Only correct egregious failures in search queries, e.g. searching for a specification not mentioned in the instruction. Otherwise search usually does not require any correction. 

(5) Considerations for click.
* When suggesting corrected_action, make sure this action exists in the list of candidate_actions at that timestep.

### Output
The output is a JSON containing a summary and a trajectory with the same length as the input student trajectory as follows:
```json
{
    "summary": your summary of the mistakes the student is making,
    "trajectory": [
    {
        "timestep": Index of the current timestep,
        "is_corrected": True/False depending on whether this timestep is corrected or not
        "corrected_reason": The corrected reason that the student should generate. If is_corrected=False, copy over original reason. ,
        "corrected_action": The corrected action that the student should take (chosen from list of candidate_actions at timestep t). If is_corrected=False, copy over original action. If click action, make sure corrected_action belongs to list of candidate_actions at the corresponding timestep in input student_trajectory. 
    },
    {
        ...
    }
    ...
    ]
}
```
{% endif %}
{% if not system %}
The student trajectory is below:
{{student_trajectory}}

The privileged state for the task is below:
{{privileged_state}}

Provide the ### Output in the JSON format specified above.
{% endif %}
\end{lstlisting}

\subsubsection{Student agent template}
\begin{lstlisting}[language=Python]
{% if mode == 'input' %}
You are an intelligent shopping agent. Your objective is to solve shopping tasks by searching for items in a website, clicking on links until you solve the task.

Below is the history of previous observations and actions:
{{ observation_action_history }}

You are given as input the current observation that contains instructions for the task and contents of the current webpage. You are also given a list of possible candidate_actions:
{
    "observation": {{observation}},
    "candidate_actions": {{candidate_actions}}
}

Your goal is to generate the action to take at this time step along with the reason for taking the action. 

Please follow these general instructions:
* You MUST choose action from the list of candidate_actions. If candidate_actions has only click actions, you can only choose a click action from the list. If candidate_actions has only search[<search query>], you must search by generating the search keywords on your own.
* When choosing a search action, you MUST follow the format search[<search query>]
* Choose search keywords that are informative but not too specific or too broad. Examples of search: 
search[6 foot coaxial cable, pack of 3]
search[satin brass frosted hallway light fixture]
* When choosing click action, make sure action is VERBATIM copied from the list of candidate_actions. Examples of click:
click[satin brass | frosted]
click[buy now]
* Often times the task requires you to find objects not in the current webpage. You must search the webpages to locate the object and verify it indeed matches the specifications in the instructions. If not, go back and refine your search.
* Consult the history of previous observations and actions to see what actions you have tried already so as to not repeat your actions.
* Do NOT repeat the same action as the last action in your observation_action_history. It's going to yield the same result. 
* When choosing click action, make sure action is VERBATIM copied from the list of candidate_actions.
* Important: You must try to finish the task and click on buy now as quickly as possible. It's better to purchase a suboptimal item than not finish the task within the time horizon. You get partial points for matching a subset of the criteria in the instruction. You get 0 points for not finishing the task in the time horizon (typically 20 timesteps).

You need to generate a response in the following format. Please issue only a single action at a time.
REASON:
Rationale for what action to take next based on the task and previous history. In your reason, consult candidate_actions to precisely state VERBATIM which action you will do. 
ACTION:
The action to be taken, chosen ONLY from candidate_actions
{% elif mode == 'output' %}
REASON:
{{ reason }}
ACTION:
{{ action }}
{% endif %}
\end{lstlisting}

\subsubsection{Autolabel reasons template} 

\begin{lstlisting}[language=Python]
{% if system %}
You are given a trajectory containing observations and actions generated by an agent solving a web shopping task. Your goal is to generate the reasoning process that the agent must have been thinking at each timestep to justify each of their actions.

### Input
You will be provided with a JSON file that logs the agent's observation and action at every timestep while solving the web shop task. 

The structure is an array of objects, containing the following at each timestep:
- timestep: Index of the current timestep
- observation: The instruction the agent has to solve and the current webpage observed by the agent at a specific timestep.
- action: The action taken by the agent.


### Task
* Analyze the agent trajectory 
* Generate reason for the agent at every timestep. The reason should be in first person, e.g. "I should .."
* Base your generated reason solely on the historical observations and actions up to each timestep
* Offer GENERAL principles or hints in your reasons that explain why the agent took the action
* For example, if the agent is backtracking by clicking on previous, explain why that is the case
* Do NOT include any information from future observations as the agent does not know the future at that time
* Ensure the output maintains the same sequence of actions as in the input.
* In your REASON, be sure to provide detailed justification for the action. If candidate_actions has only click actions, specify which option from the list are you choosing as your action.
* Often times, the task requires you to select both a product and specific options for that product. Be sure to justify why you are selecting a particular option for a product by referring to the task in your observation.
* Often times, the agent settles for an option even though it may not be exactly optimal. This is because the agent has a fixed time horizon within which they must click buy now. Ensure that your reason rationalizes why it's good to finish the task by clicking on buy now. 

### Output
The output is a trajectory represented as a JSON array of objects with the same length as the input trajectory as follows:
```json
[
    {
        "timestep": Index of the current timestep,
        "reason": A reason to justify the action given observation and action up until the current timestep
        "action": The action taken by the agent (copy as is from input trajectory)
    },
    ...
]
{% endif %}
{% if not system %}
The input trajectory is below:
{{input}}

Provide the ### Output in the JSON format specified above.
{% endif %}
\end{lstlisting}

\subsection{InterCode}
\label{sec:appendix:prompts:intercode}

\subsubsection{{Expert corrections template}}
\begin{lstlisting}[language=Python]
{% if system %}
You are given a trajectory containing observations, reason and actions generated by a student agent interacting with a MySQL Database system using sql queries to answer a question.  You are a teacher who has access to a "privileged state" that contains secret information sufficient to solve the task but is hidden from the student. Your goal is to improve how the student solves the task by improving their reason and action at every timestep. 

### Input
You will be provided with a JSON file that logs the student agent's observation, reason and action at every timestep while solving a shopping task. The student fails to solve the task within the time horizon of the task. 

The structure is an array of objects, containing the following at each timestep:
- timestep: Index of the current timestep
- observation: The observation provided to the student agent
- reason: The reason generated by the student agent to justify their action
- action: The action taken by the student agent

You will also be provided the privileged state. The privileged state contains an example of an optimal SQL command to solve the task and the expected observation that solves the task. Note that it is the observation that the student gets evaluated on. Multiple possible actions can lead to that observation.

### Task
* Analyze the student trajectory and summarize the main mistakes the student is making when trying to solve the task.
* Refer to your privileged state to understand how the task can be solved optimally.
* Correct a select set of timesteps where the student's action is clearly wrong.
* Generate **IMPROVED REASONS** and **ACTIONS** for the student at each timestep to guide them towards the goal. The improved reason should be framed as if it is the student's new reasoning, e.g., "I should ...".
* The improved reason and action are intended to replace the student's original reason and action at that timestep. Do NOT refer to the original incorrect reason and action when generating improved reason and action. 
* Base your improved reasons solely on the student's historical observations and actions up to that timestep.
* Do NOT include any information from your privileged state in the improved reasons, as the student does not have access to that information.
* Offer general principles or hints in your improved reasons that explain why the student should prefer your suggested action over their original action. This would help the student generalize better.
* When generating an improved reason and action at timestep `t`, assume the student followed their original trajectory up until timestep `t`.

Important:
(1) Provide **GENERAL principles or hints** in your improved reasons.
* If you know from your privileged state that a particular set of tables should be queried to solve the task, suggest reasons and actions that guide the student to first discover those tables.
* Do not directly instruct the student to output the privileged action, as they do not have access to privileged information.
* By following these steps, you help the student understand the logic behind the actions without revealing privileged information.

(2) Apply corrections sparingly.
* Be strategic about which timesteps you choose to correct.


### Output
The output is a JSON containing a summary and a trajectory with the same length as the input student trajectory as follows:
```json
{
    "summary": your summary of the mistakes the student is making,
    "trajectory": [
    {
        "timestep": Index of the current timestep,
        "original_reason": The original reason generated by the student at timestep t (copy as is from student trajectory),
        "original_action": The original action the student took at timestep t (copy as is from student trajectory),
        "is_corrected": True/False depending on whether this timestep is corrected or not,
        "corrected_reason": The corrected reason that the student should generate at timestep t. If is_corrected=False, leave blank ,
        "corrected_action": The corrected action that the student should take at timestep t. If is_corrected=False, leave blank 
    },
    {
        ...
    }
    ...
    ]
}
```
{% endif %}
{% if not system %}
The student trajectory is below:
{{student_trajectory}}

The privileged state for the task is below:
{{privileged_state}}

Provide the ### Output in the JSON format specified above.
{% endif %}
\end{lstlisting}

\subsubsection{Student agent template}
\begin{lstlisting}[language=Python]
{% if system %}
You are given a trajectory containing observations, reason and actions generated by a student agent interacting with a MySQL Database system using sql queries to answer a question.  You are a teacher who has access to a "privileged state" that contains secret information sufficient to solve the task but is hidden from the student. Your goal is to improve how the student solves the task by improving their reason and action at every timestep. 

### Input
You will be provided with a JSON file that logs the student agent's observation, reason and action at every timestep while solving a shopping task. The student fails to solve the task within the time horizon of the task. 

The structure is an array of objects, containing the following at each timestep:
- timestep: Index of the current timestep
- observation: The observation provided to the student agent
- reason: The reason generated by the student agent to justify their action
- action: The action taken by the student agent

You will also be provided the privileged state. The privileged state contains an example of an optimal SQL command to solve the task and the expected observation that solves the task. Note that it is the observation that the student gets evaluated on. Multiple possible actions can lead to that observation.

### Task
* Analyze the student trajectory and summarize the main mistakes the student is making when trying to solve the task.
* Refer to your privileged state to understand how the task can be solved optimally.
* Correct a select set of timesteps where the student's action is clearly wrong.
* Generate **IMPROVED REASONS** and **ACTIONS** for the student at each timestep to guide them towards the goal. The improved reason should be framed as if it is the student's new reasoning, e.g., "I should ...".
* The improved reason and action are intended to replace the student's original reason and action at that timestep. Do NOT refer to the original incorrect reason and action when generating improved reason and action. 
* Base your improved reasons solely on the student's historical observations and actions up to that timestep.
* Do NOT include any information from your privileged state in the improved reasons, as the student does not have access to that information.
* Offer general principles or hints in your improved reasons that explain why the student should prefer your suggested action over their original action. This would help the student generalize better.
* When generating an improved reason and action at timestep `t`, assume the student followed their original trajectory up until timestep `t`.

Important:
(1) Provide **GENERAL principles or hints** in your improved reasons.
* If you know from your privileged state that a particular set of tables should be queried to solve the task, suggest reasons and actions that guide the student to first discover those tables.
* Do not directly instruct the student to output the privileged action, as they do not have access to privileged information.
* By following these steps, you help the student understand the logic behind the actions without revealing privileged information.

(2) Apply corrections sparingly.
* Be strategic about which timesteps you choose to correct.


### Output
The output is a JSON containing a summary and a trajectory with the same length as the input student trajectory as follows:
```json
{
    "summary": your summary of the mistakes the student is making,
    "trajectory": [
    {
        "timestep": Index of the current timestep,
        "original_reason": The original reason generated by the student at timestep t (copy as is from student trajectory),
        "original_action": The original action the student took at timestep t (copy as is from student trajectory),
        "is_corrected": True/False depending on whether this timestep is corrected or not,
        "corrected_reason": The corrected reason that the student should generate at timestep t. If is_corrected=False, leave blank ,
        "corrected_action": The corrected action that the student should take at timestep t. If is_corrected=False, leave blank 
    },
    {
        ...
    }
    ...
    ]
}
```
{% endif %}
{% if not system %}
The student trajectory is below:
{{student_trajectory}}

The privileged state for the task is below:
{{privileged_state}}

Provide the ### Output in the JSON format specified above.
{% endif %}
\end{lstlisting}

\end{document}